\documentclass[11pt]{article}
\usepackage{acl}  

\usepackage{times}
\usepackage{latexsym}
\usepackage[T1]{fontenc}
\usepackage[utf8]{inputenc}
\usepackage{microtype}
\usepackage{inconsolata}
\usepackage{graphicx}
\usepackage{booktabs}
\usepackage{amsmath}
\usepackage{amssymb}  
\usepackage{multirow}
\usepackage{array}
\usepackage{tabularx}
\usepackage{subcaption}
\usepackage{enumitem}
\usepackage{algorithm}
\usepackage{algpseudocode}
\usepackage{adjustbox}
\usepackage{xcolor}
\usepackage{listings}
\lstdefinestyle{prompt}{
  basicstyle=\ttfamily\footnotesize,
  breaklines=true, breakatwhitespace=true,
  columns=fullflexible, frame=single, framesep=4pt,
  xleftmargin=4pt, xrightmargin=4pt, upquote=true,
  literate={->}{{$\rightarrow$}}2 {—}{{---}}1 {–}{{--}}1
           {×}{{$\times$}}1 {≈}{{$\approx$}}1 {≥}{{$\geq$}}1
}
\lstdefinestyle{promptstyle}{
  basicstyle=\ttfamily\scriptsize,
  breaklines=true, breakindent=0pt, columns=fullflexible,
  keepspaces=true, showstringspaces=false,
  frame=single, framesep=4pt, rulecolor=\color{gray!50},
  backgroundcolor=\color{gray!5},
  literate={—}{{-{}-}}1 {–}{{-}}1 {×}{{$\times$}}1 {≈}{{$\approx$}}1
           {→}{{$\rightarrow$}}1 {±}{{$\pm$}}1 {≥}{{$\geq$}}1
           {’}{{'}}1 {“}{{``}}1 {”}{{''}}1
}

\title{ACE: A Self-Correcting Agentic Canvas Editor for\\
 Multi-Slide Presentation Automation}

\author{JooYoung Jang$^{1,2}$ \quad Taegyeong Lee$^{2}$ \quad
  Jihyeon Park$^{2}$ \quad Nojun Kwak$^{1}$ \\
  $^{1}$Seoul National University \quad $^{2}$Miridih \\
  \texttt{jyjang1090@snu.ac.kr, tglee@miridih.com} \\
  \texttt{milhaud1201@gmail.com, nojunk@snu.ac.kr}}

\newcommand{\blfootnote}[1]{%
  \begingroup
  \renewcommand{\thefootnote}{}\footnote{#1}%
  \addtocounter{footnote}{-1}%
  \endgroup
}

\begin{document}
\maketitle
\blfootnote{\raggedright
\mbox{Code: {\fontsize{7}{8.5}\selectfont\href{https://github.com/BloomBerry/agentic-canvas-editor}{\nolinkurl{github.com/BloomBerry/agentic-canvas-editor}}}}\\
\mbox{Dataset: {\fontsize{7}{8.5}\selectfont\href{https://huggingface.co/datasets/BloomBerry/figma-slide-benchmark}{\nolinkurl{hf.co/datasets/BloomBerry/figma-slide-benchmark}}}}}

\begin{abstract}
Commercial design platforms increasingly edit documents through large
language model (LLM) agents, but two practical problems block reliable
deployment: legacy document formats expose only \emph{flat}, absolutely
positioned elements, so agents must recompute coordinates and routinely
break layouts; and design has no unique ground truth, so
diff-against-reference metrics penalize valid-but-different outputs.
We present \textbf{ACE}, an agentic canvas editor over a
\emph{hierarchical scene-graph} with a presentation-specialized action
space (98 tools), paired with \textbf{CARE}, a content-aware router that
feeds the agent only the relevant slice of each deck (avg.\ $\sim$89\%
input-token reduction), and a \emph{self-correction} loop driven by a
\emph{ground-truth-free} instruction-following (IF) judge whose
natural-language critique is fed back as the next-turn instruction. With
a fixed backbone, a scene-graph editor in a \emph{single turn} already
matches a same-backbone \emph{agentic} HTML pipeline that iterates
internally; adding self-correction lifts ACE significantly above it on
instruction following (IF 4.23 vs.\ 3.81 on the full 94-task benchmark,
paired $p{=}.010$, replicated by an out-of-loop judge) at 1.75$\times$
the speed and $\sim$44\% lower cost. VQ means are statistically
indistinguishable, but 26 blind raters prefer ACE overall (58.7\%
decisive win-rate) and prefer the self-corrected output 81\% of the time;
the ranking is invariant across three judge families, and out-of-loop
judges retain two-thirds of the self-correction gain, bounding
circularity. 66\% of cases halt after one pass, and a strict-peak
rollback removes every observed regression.
\end{abstract}

\begin{figure*}[t]
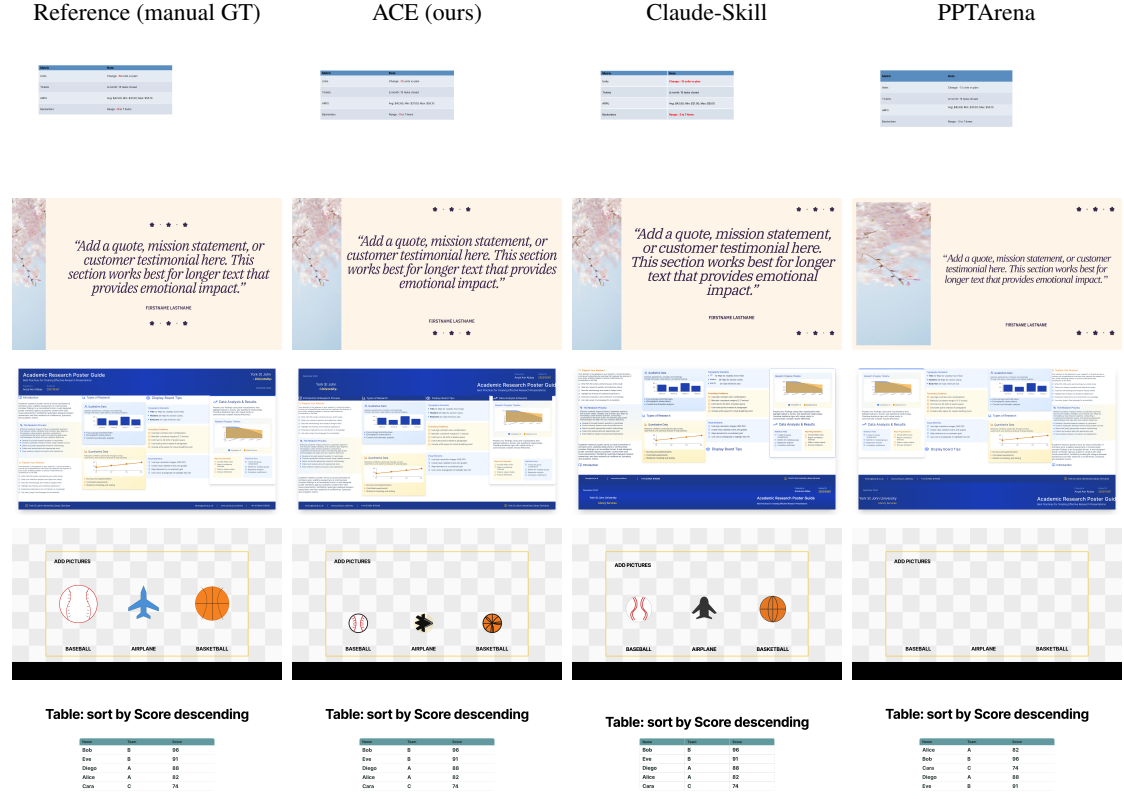

  \centering
  \setlength{\tabcolsep}{2pt}
  \renewcommand{\arraystretch}{1.0}
  
  \let\myincludegraphics\includegraphics
    \newcommand{\acell}[1]{\myincludegraphics[width=0.235\textwidth,height=2cm,keepaspectratio]{latex/images/#1}}
  \newcommand{\nores}{\colorbox{gray!12}{\parbox[c][1.85cm][c]{0.205\textwidth}{\centering\scriptsize\itshape no valid\\ output}}}
  \begin{tabular}{@{}cccc@{}}
    \footnotesize Reference (manual GT) & \footnotesize ACE (ours)
      & \footnotesize Claude-Skill & \footnotesize PPTArena \\[1pt]
    \acell{qual_case71_ref.png} & \acell{qual_case71_ours.png}
      & \acell{qual_case71_html.png} & \acell{qual_case71_pptarena.png} \\[1pt]
    \acell{qual_case32_ref.png} & \acell{qual_case32_ours.png}
      & \acell{qual_case32_html.png} & \acell{qual_case32_pptarena.png} \\[1pt]
    \acell{qual_case42_ref.png} & \acell{qual_case42_ours.png}
      & \acell{qual_case42_html.png} & \acell{qual_case42_pptarena.png} \\[1pt]
    \acell{qual_case56_ref.png} & \acell{qual_case56_ours.png}
      & \acell{qual_case56_html.png} & \acell{qual_case56_pptarena.png} \\[1pt]
    \acell{qual_case61_ref.png} & \acell{qual_case61_ours.png}
      & \acell{qual_case61_html.png} & \acell{qual_case61_pptarena.png} \\
  \end{tabular}
  \caption{Qualitative comparisons (each row: reference, ACE, Claude-Skill
HTML,
    PPTArena). \textbf{(a)}~Case~71, highlight only negative table cells;
    \textbf{(b)}~Case~32, arrange image and text;
    \textbf{(c)}~Case~42, organize a research poster;
    \textbf{(d)}~Case~56, add pictures;
    \textbf{(e)}~Case~61, sort rows by score and crop images to 16:9.
    Backbones: \texttt{claude-sonnet-4-6} for ACE/HTML; \texttt{gpt-5.5} judge.
    Because design has no unique ground truth, ACE's edits often differ
    from the manual reference yet remain valid---what our reference-free
    IF judge rewards.}
  \label{fig:qual_appendix}
\end{figure*}

\section{Introduction}

Most presentations begin from a \emph{template}, not a blank canvas.
Software such as Microsoft PowerPoint and Canva is used at scale precisely
because starting from a professionally designed template yields
higher-quality results with far less effort---the user inherits an expert
layout and edits only what matters~\citep{nouraei2024thinking}. Editing that template, however, is
itself laborious: reflowing elements, matching a theme, turning a bullet
list into a diagram, or rebuilding a chart all demand repetitive,
detail-sensitive manual work~\citep{jung2025talktoyourslides}.

A growing body of work automates this with LLM agents. PPTArena
\citep{pptarena} and PPTAgent \citep{pptagent} edit existing PowerPoint
decks; AutoPresent \citep{autopresent} and SlideCoder \citep{slidecoder}
synthesize slides as executable programs; and AeSlides \citep{aeslides}
optimizes layout aesthetics with verifiable rewards. These works are
valuable, but two problems block their use for template editing.
\textbf{(P1) Design has no unique ground truth.} PPTArena
\citep{pptarena} scores predictions against a single reference under a
fixed \texttt{style\_target}---a per-sample rubric distilled from that
one reference deck---penalizing different-but-valid edits, while
SlideCoder \citep{slidecoder} and AutoPresent \citep{autopresent} turn a
reference \emph{image} into an editable slide rather than editing an
existing deck. Judging whether the \emph{user's intent} was met---and
iterating until it is---matches how designers actually work
\citep{uifeedback, revisionmatters}.
\textbf{(P2) Code-generation editing does not scale, and previous editable
action spaces are too small.} Emitting \texttt{python-pptx} leans entirely
on the model's raw capability and absolute-coordinate arithmetic, which
breaks down on complex, multi-slide decks; and structured alternatives
expose too few operations---PPTAgent \citep{pptagent} offers only five
editable APIs---to cover the range of real-world edits, so quality degrades
as decks grow.

We address both with \textbf{ACE}, a self-correcting agentic canvas
editor, and make three contributions:
\begin{enumerate}[noitemsep,topsep=2pt,leftmargin=*]
\item A \textbf{reference-free, edit-grounded multimodal
instruction-following evaluator} (for P1) that scores the agent's
\emph{initial$\to$prediction} edit delta against the instruction---rather
than against a single ground-truth answer---and selectively attaches
rendered images to verify visual-semantic intent. Its score and critique
drive a self-correction loop. Unlike ReAct- and Reflexion-style loops
\citep{react,reflexion}, which presuppose an environment-supplied verify
signal (an execution error, a task reward), open-ended design editing
offers no natural success signal; we show a rendered-diff IF critique is a
usable \emph{GT-free reward} there, and validate it with a 26-rater blind
human study (76--80\% agreement on decided cases) and two out-of-loop
judge families that preserve every headline ranking
(\S\ref{sec:judgevalid}).
\item A \textbf{domain-specific action space} (for P2) over a scene-graph
that is hierarchical like the HTML DOM yet editable like OOXML, paired
with a content-aware context router (CARE); together they improve cost
and speed over code-generation editing, and leave-one-out ablations at a
fixed backbone and judge show the representation, the router, the
specialized tools, and self-correction each contribute independently
(\S\ref{sec:abl}).
\item A \textbf{benchmark of 97 multi-slide tasks (94 evaluable)} for the
Figma-Slides setting (12 novel, 9 with no PowerPoint analogue), released
with execution logs and analysis scripts, on which ACE leads an
off-the-shelf coding agent paired with a Claude Skill and an OpenXML
baseline on \textbf{instruction following (IF)}, significantly under two
independent judges (\S\ref{sec:stats}); visual
quality (VQ) means are indistinguishable from the strongest baseline but
resolved in ACE's favor by blind human preference
(\S\ref{sec:judgevalid}).
\end{enumerate}

\section{Related Work}
\label{sec:related}

\subsection{Presentation Editing}
\label{sec:rw-edit}
Because slide decks are multi-slide artifacts, editing must preserve both
narrative and visual consistency across pages. PPTArena \citep{pptarena}
edits PowerPoint via code generation and XML patching, while PPTAgent
\citep{pptagent} edits through a small set of five editable operations;
both are cost-efficient but perform poorly on complex edits. More
recently, Claude Skills\footnote{\url{https://claude.com/skills}}, defined
by a \texttt{SKILL.md} specification, treat slides as LLM-friendly HTML and
edit them with generic read/edit/bash operations: reading the entire deck
yields consistent output, but at high cost and latency. The former family
is thus cheap but weak, while the latter is consistent but expensive. We
instead propose CARE, a content-aware router that extracts only the deck
context each instruction needs (\S\ref{sec:care}), so that ACE outperforms
a commercial coding agent (Claude Code) paired with an Agent Skill on
instruction following---with blind human raters preferring ACE overall
(\S\ref{sec:judgevalid})---while running faster and at lower cost.

\subsection{Self-Correction with Verifier Feedback}
\label{sec:rw-verify}
A growing body of work improves generation by closing a
\emph{propose--verify--refine} loop. VASCAR \citep{vascar} performs content-aware layout
generation with a frozen LVLM, turning geometric metrics (overlap,
occlusion) into a textual instruction re-applied for several rounds. In
the slide domain the verify signal is typically an execution error or a
render: SlideCoder \citep{slidecoder} re-prompts with the captured error
and an API grammar, PPTAgent \citep{pptagent} repairs REPL error logs,
and AutoPresent \citep{autopresent} feeds a rendered snapshot back to fix
spacing and placement. Re-rendering the full deck at every iteration, however, is
costly and can miss fine edits that are hard to discern visually. PPTArena
\citep{pptarena} adopts a ReAct-style \citep{react} loop that re-feeds
rendered screenshots of the changed slides, and leaves an explicit in-loop
judge to future work. We adopt the same closed-loop skeleton but contribute
a verify signal tailored to instruction-grounded editing: a reference-free,
multimodal IF verifier over the agent's structural edit diff that scores
the original$\to$prediction delta and attaches images only when needed
(\S\ref{sec:selfcorrect}). Where ReAct \citep{react} and Reflexion
\citep{reflexion} presuppose that a verify signal already exists---an
execution error, a task reward, environment feedback---open-ended design
editing offers none; the GT-free reward is what turns this unverifiable
task into a verifiable loop.

\subsection{Tool-Augmented LLMs}
\label{sec:rw-tool}
Toolformer \citep{toolformer} showed that LLMs can extend their abilities
through external tools, prompting a wave of tool-use research. Canvas
\citep{canvas} provides a 52-tool action space for vision-language agents
on Figma \emph{Design}, and PPTAgent \citep{pptagent} defines five tools
for basic delete/copy/replace edits. Unlike these, we target a scene-graph
that is hierarchical like the HTML DOM yet editable like XML, and provide
98 tools specialized to it (\S\ref{sec:ace}), yielding performance
advantages.

\section{Self-Correcting Agentic Canvas Editor}
\label{sec:system}
\begin{figure*}[t]
\centering
\includegraphics[width=\textwidth]{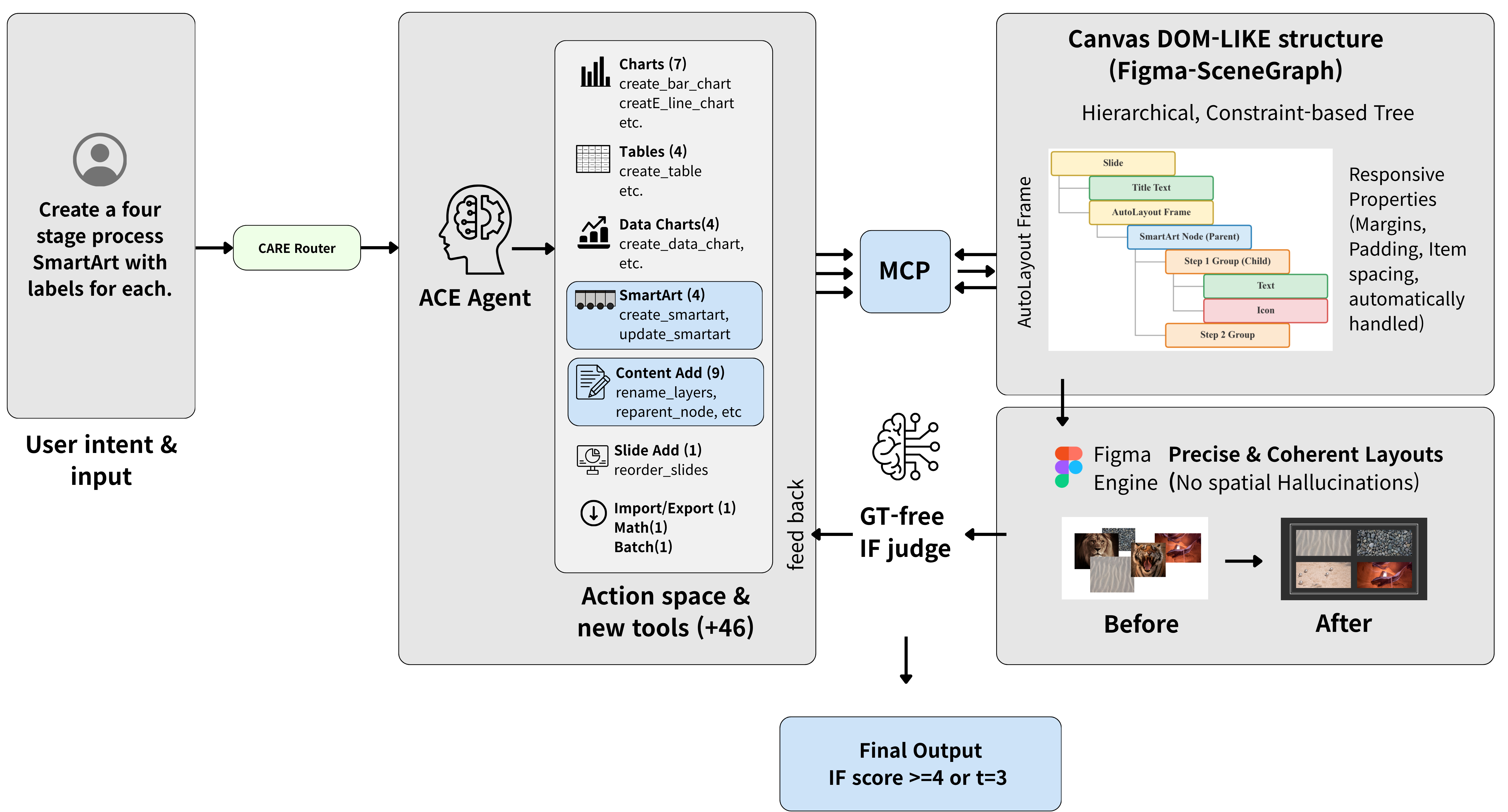}
\caption{Overall architecture of ACE: the scene-graph editor and its
98-tool action space, CARE's three-way context routing, and the
self-correction loop driven by the ground-truth-free IF judge.}
\label{fig:ace_overview}
\end{figure*}

Figure~\ref{fig:ace_overview} gives the
overall closed-loop architecture; we describe each component below.

\subsection{Scene-Graph Domain-Specific Action Space}
\label{sec:ace}

\begin{figure*}[t]
\centering
\begin{subfigure}[b]{0.30\textwidth}
  \centering
  \includegraphics[height=2.8cm,keepaspectratio]{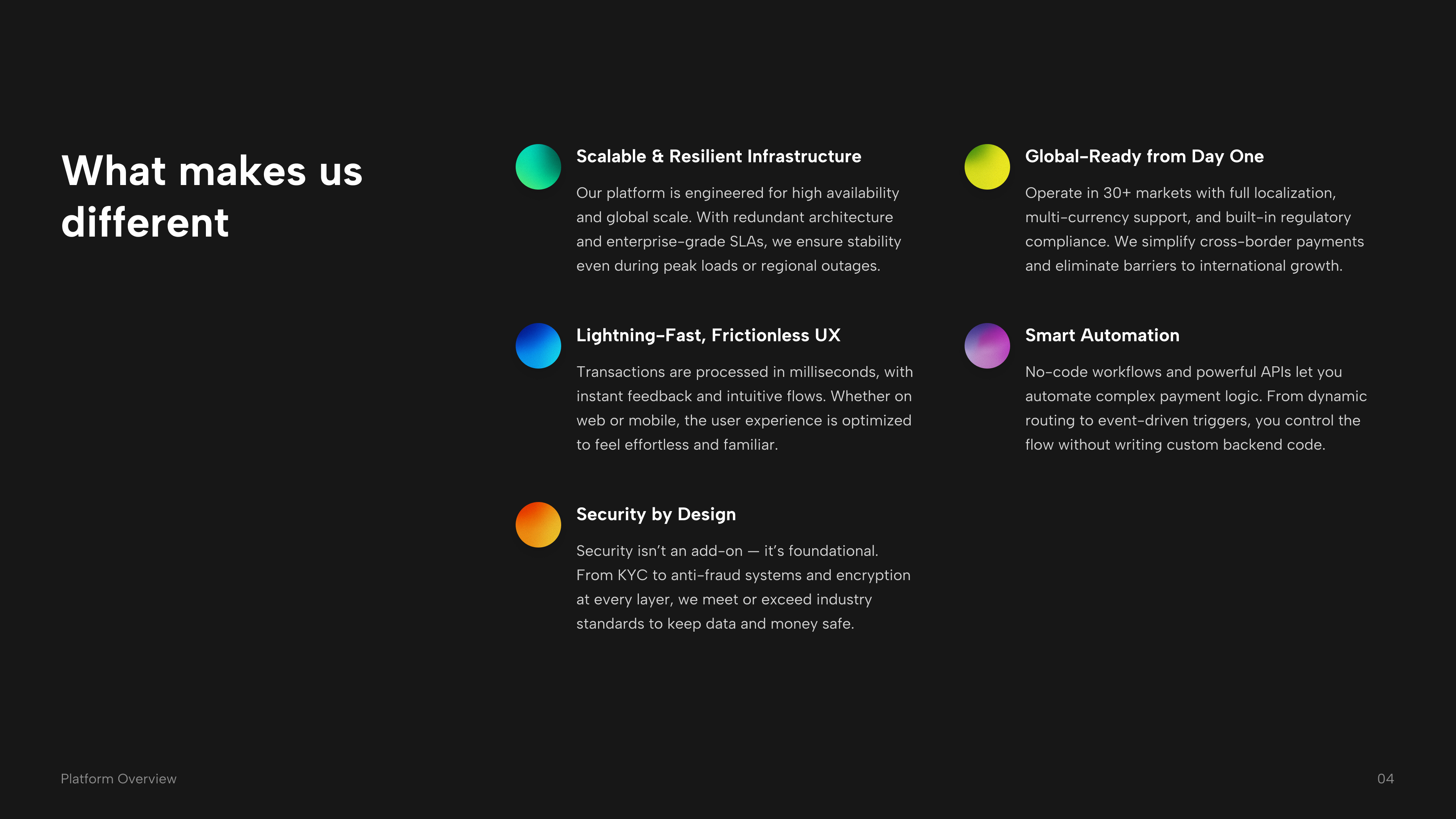}
  \caption{Before.}
  \label{fig:ds_before}
\end{subfigure}
\hfill
\begin{subfigure}[b]{0.30\textwidth}
  \centering
  \includegraphics[height=2.8cm,keepaspectratio]{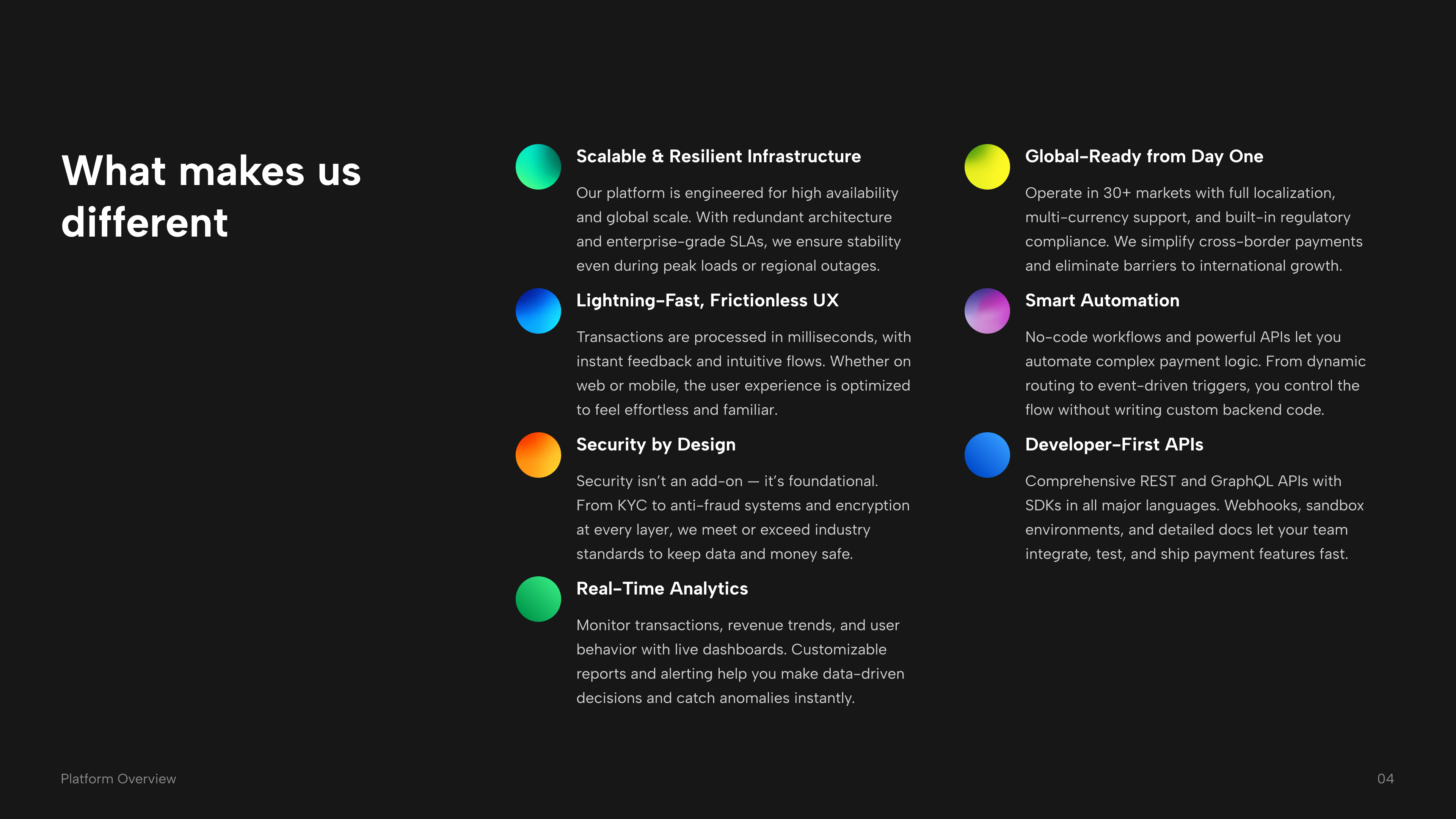}
  \caption{After (auto-layout $5\to7$).}
  \label{fig:ds_after}
\end{subfigure}
\hfill
\begin{subfigure}[b]{0.36\textwidth}
  \centering
  \includegraphics[height=2.8cm,keepaspectratio]{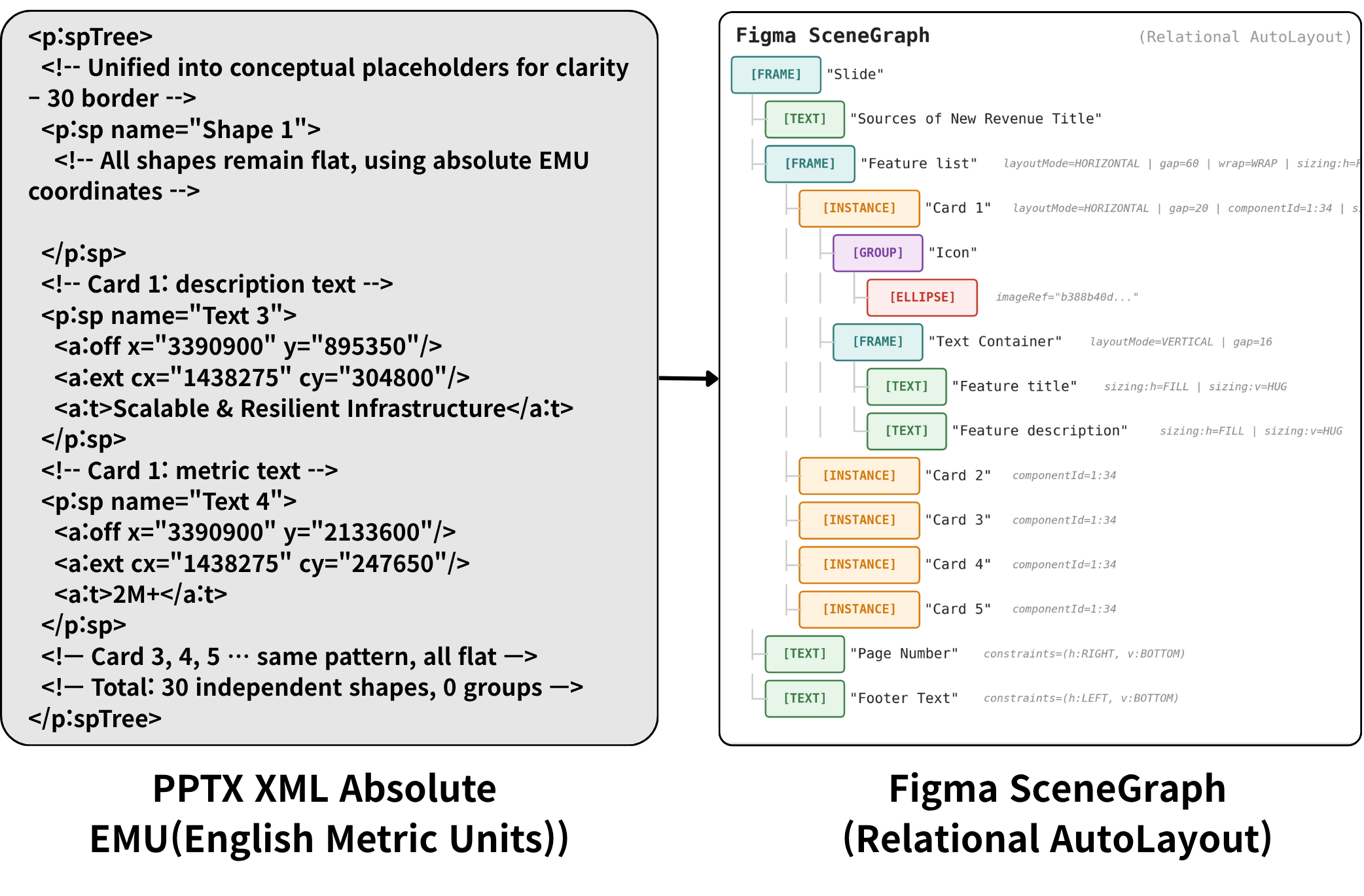}
  \caption{OOXML vs.\ Figma SceneGraph.}
  \label{fig:ds_repr}
\end{subfigure}
\caption{A slide, an example edit, and its data representations.
(a,b) Case~112 increases the auto-layout children from five to seven and
the frame re-flows automatically using auto-layout. (c) OOXML flattens the slide into
absolutely positioned siblings (EMU coordinates, no grouping), while the
SceneGraph encodes it as a parent--child tree with responsive auto-layout
constraints.}
\label{fig:datastructure}
\end{figure*}

\paragraph{The legacy format and its limit.}
PowerPoint's Office Open XML (OOXML) stores a slide as a \emph{flat}
sequence of shapes under \texttt{<p:spTree>}, each pinned by an
\emph{absolute} offset in English Metric Units---e.g.\
\texttt{<a:off x="3390900"/>} places a shape 3{,}390{,}900\,EMU
($\approx$3.7\,in) from the slide origin
(Figure~\ref{fig:datastructure}c). Because there is no hierarchy or
grouping, editing one element forces the agent to recompute the
coordinates of every other shape to avoid overlaps---a
``coordinate-pushing'' regime that produces spatial hallucinations
whenever content is inserted or removed.

\paragraph{Our solution: a scene-graph.}
ACE follows the scene-graph used by Figma Slides~\citep{FigmaSlidesDoc} and reads a slide as a
\textbf{parent--child tree}. Each child is placed by a transform
\emph{relative} to its parent, so moving or resizing the parent updates all
of its descendants automatically through the applied constraints. A parent
may optionally enable \emph{auto-layout}~\citep{figmaautolayout}, which fixes child placement by
logic rather than numbers: a child can \texttt{fill} its parent, a parent
can \texttt{hug} its children, or a dimension can be \texttt{fixed}
independently of either. For example, given the instruction \emph{``increase
the auto-layout children in slide 4 from five to seven and fill in the
contents''} (Case~112): in a flat format this requires repositioning all
five existing cards; under auto-layout the agent inserts the two new
children and the parent re-flows the row
(Figure~\ref{fig:datastructure}a,b), never computing per-child
coordinates.

\paragraph{Action space.}
On this representation we extend the 52-tool Canvas UI action space
\citep{canvas} to a \textbf{98-tool} suite specialized for multi-slide
editing, organized into 11 modules (Appendix~\ref{app:tools}). The key
choice is to expose \emph{semantic} operations (e.g.\
\texttt{create\_smartart}, \texttt{create\_data\_chart}) rather than
primitive shape calls, so the agent maps intent directly to structured
entities: a table-to-chart conversion that takes 66 primitive operations
collapses to 22 with the specialized \texttt{create\_graphics} tool (a
3.0$\times$ reduction in the execution trace; Appendix~\ref{app:tooleff}),
reducing the ``reasoning tax'' and the chance of alignment errors.

\subsection{CARE: Content-Aware Context Routing}
\label{sec:care}

\begin{figure}[t]
\centering
\includegraphics[width=\linewidth]{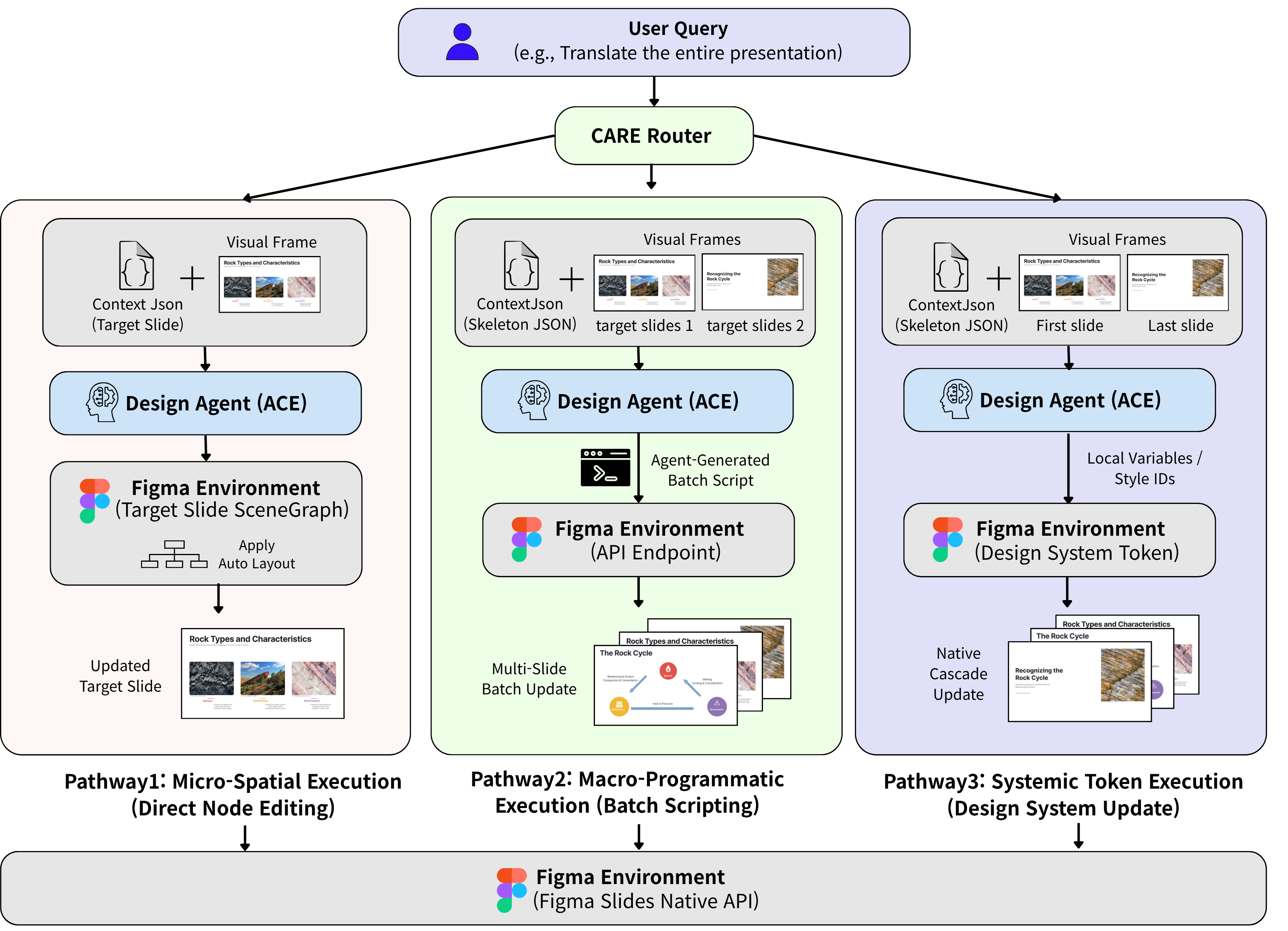}
\caption{The ACE workflow with CARE. CARE routes each instruction into one
of three context scopes---Micro-Spatial (direct node editing),
Macro-Programmatic (skeleton JSON for cross-slide batch scripting), and
Systemic-Token (style metadata for design-system updates)---minimizing the
agent's context window load while preserving precision.}
\label{fig:CAREworkflow}
\end{figure}

A full serialization of a long deck can exceed 1M tokens, overflowing
the context window. Rather than feed the agent the whole deck, CARE routes each instruction to the minimal
context it needs (Figure~\ref{fig:CAREworkflow}). A single
\emph{mode-and-target} classifier selects one of three scopes and the
target slide indices: \textbf{Micro-Spatial} (high-fidelity JSON for the
target slide only), \textbf{Macro-Programmatic} (a skeleton JSON of node
IDs and text for cross-slide batch edits), and \textbf{Systemic-Token}
(only design-system variables and style IDs). Relative to passing the full
deck, this cuts input tokens by 86.6--95.7\% on average across modes
(Table~\ref{tab:care_performance}; up to 99.9\%), keeping the agent's
context lean regardless of deck size. CARE is a cost and scalability
mechanism that does not trade off quality (full-context ablation and
routing audit: \S\ref{sec:abl}, Appendix~\ref{app:careaudit}). Routing pseudocode and the full control
flow are in Appendix~\ref{app:router}.

\begin{table}[t]
\centering\small
\adjustbox{max width=\columnwidth}{%
\begin{tabular}{@{}lcccc@{}}
\toprule
\textbf{Mode} & \textbf{N} & \textbf{Avg.\ Red.} & \textbf{Min} & \textbf{Max} \\
\midrule
Systemic-Token     & 7  & 95.7\% & 90.7\% & 99.4\% \\
Macro-Programmatic & 37 & 90.9\% & 73.3\% & 98.8\% \\
Micro-Spatial      & 50 & 86.6\% & 70.1\% & 99.9\% \\
\bottomrule
\end{tabular}}
\caption{Input-token reduction by CARE mode, relative to passing the full
deck (94 evaluable tasks).}
\label{tab:care_performance}
\end{table}

\subsection{Self-Correction with a Ground-Truth-Free IF Judge}
\label{sec:selfcorrect}

\begin{figure}[t]
\centering
\includegraphics[width=\linewidth]{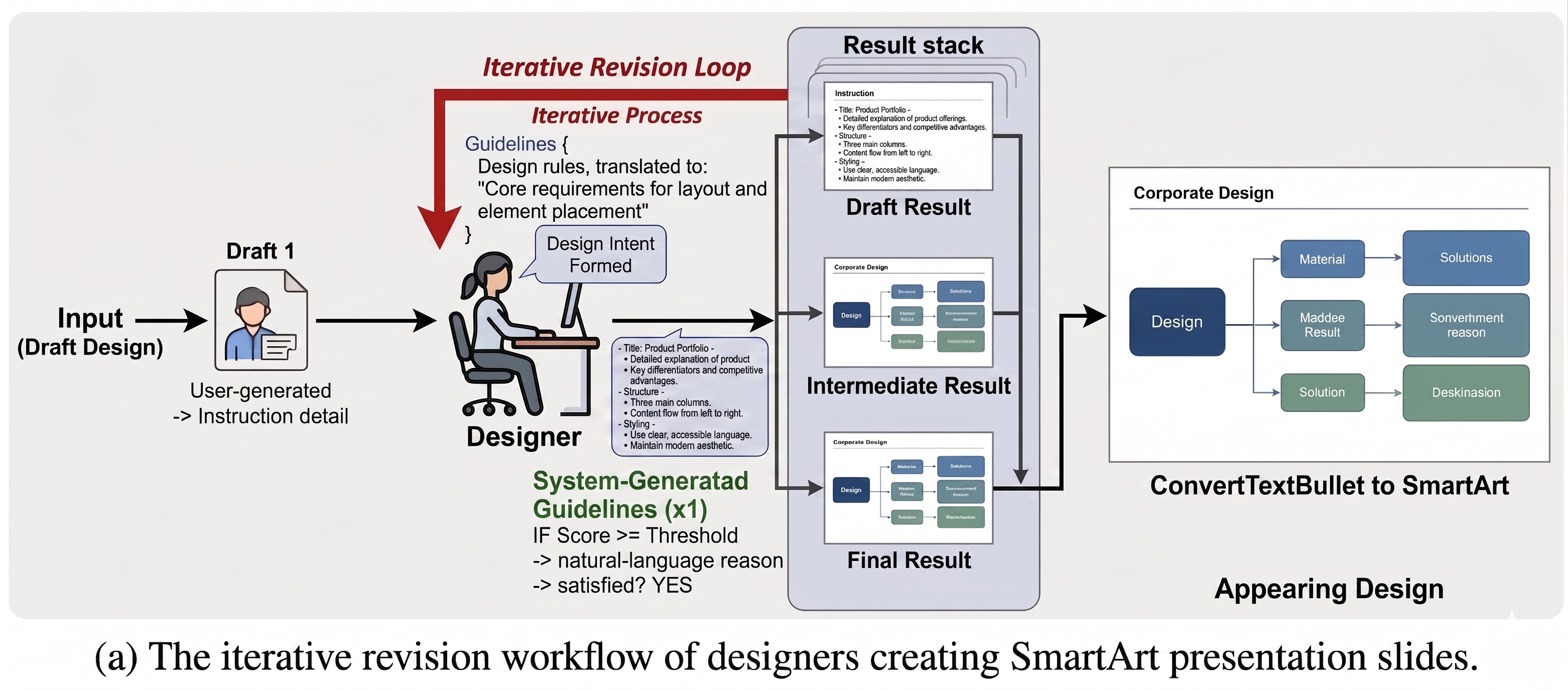}
\caption{The human designer's iterative workflow that ACE imitates: form a
design intent, edit the original, inspect the result against that intent,
and repeat until satisfied.}
\label{fig:human_workflow}
\end{figure}

\begin{figure*}[t]
\centering
\footnotesize
\setlength{\tabcolsep}{4pt}
\newcommand{\scell}[1]{\includegraphics[width=0.235\textwidth,height=2.2cm,keepaspectratio]{latex/images/#1}}
\begin{tabular}{@{}cccc@{}}
\scell{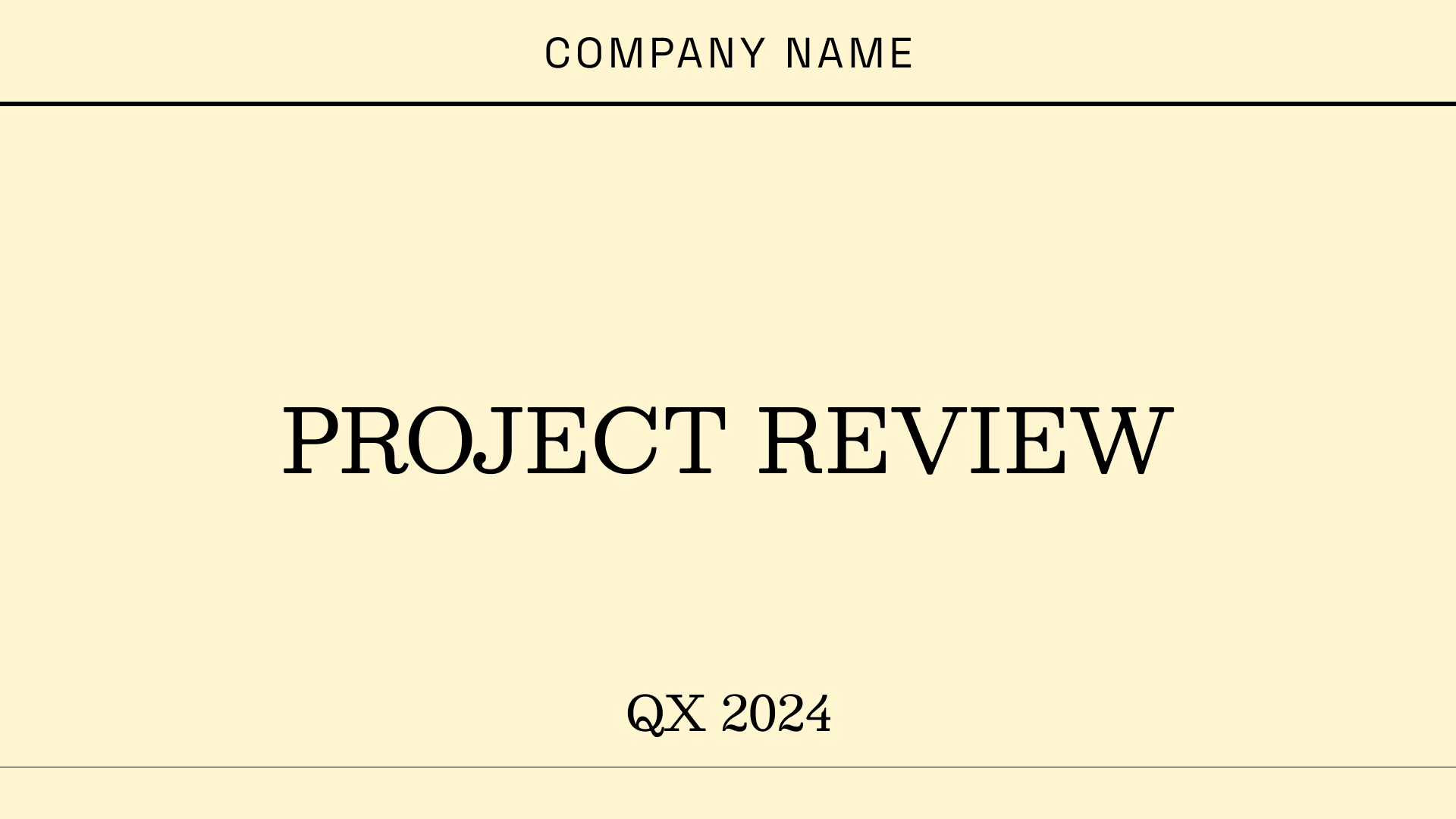} & \scell{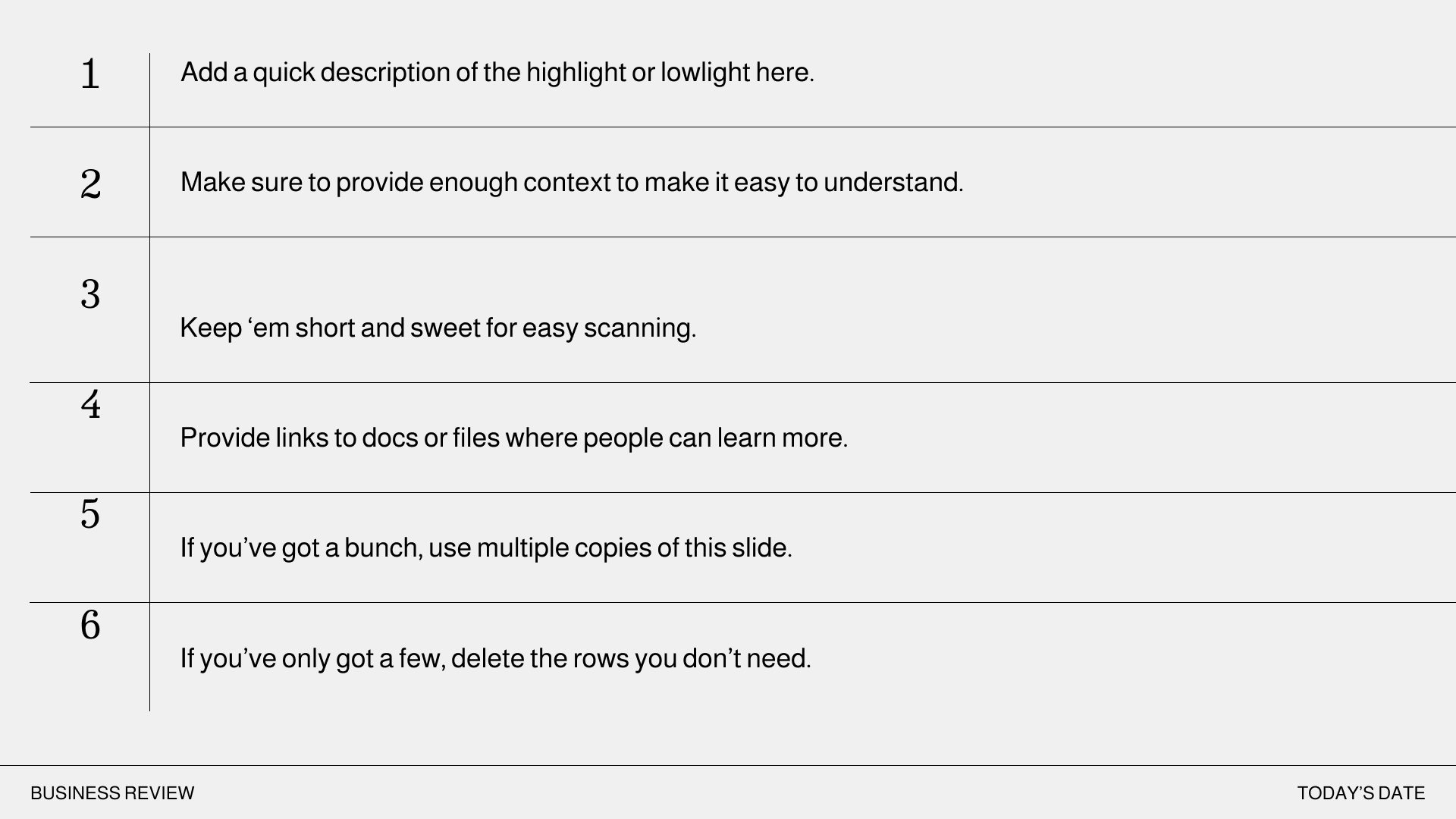}
  & \scell{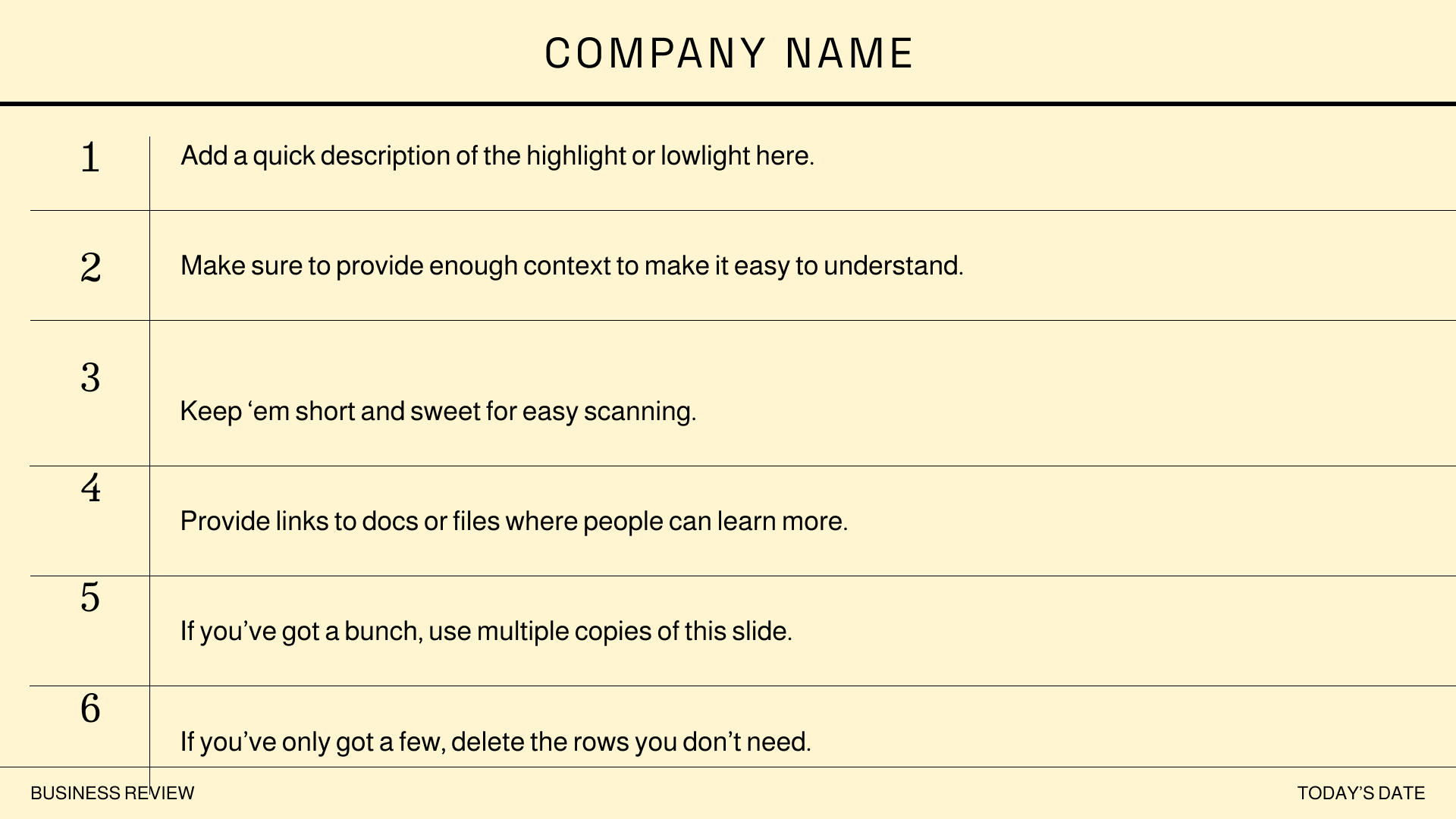} & \scell{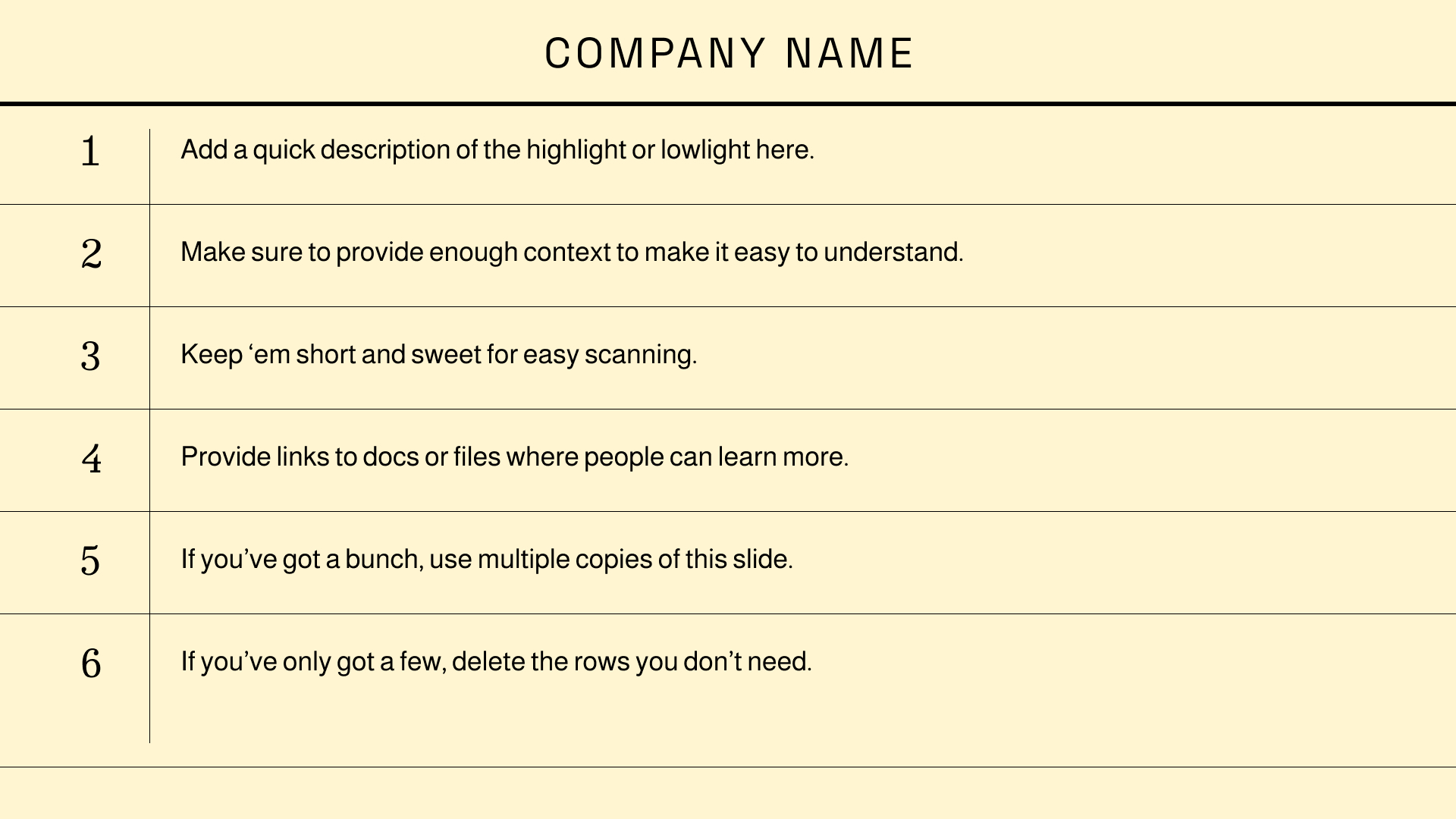} \\[2pt]
\textbf{Slide 1 (reference)} & \textbf{Origin}
  & \textbf{Iteration 1} & \textbf{Iteration 2 (final)} \\
\end{tabular}
\caption{Self-correction on Case~54 (\texttt{gpt-5.5} backbone).
\textbf{Slide~1 (reference):} the target design---cream background, centred
company header, and horizontal rules---to propagate to slides~2--9.
\textbf{Origin:} the rest of the deck before editing.
\textbf{Iteration~1 IF score = 3 (IF critique):} the agent applied some key slide~1 layout
elements to slides~2--9, including the cream background, centred company
header, and horizontal rules, and shifted some content down, but it did not
fully apply slide~1's layout structure, left old footer text/elements, and
some adjusted content does not fit cleanly within the new layout.
\textbf{Iteration~2 IF score = 4 (final):} this critique becomes the next-turn instruction;
the agent removes the residual footer elements and refits the displaced
content, yielding a layout consistent with slide~1 across the deck---an
instance of the loop in Algorithm~\ref{alg:selfcorrect}.}
\label{fig:overall}
\end{figure*}

Template editors rarely edit blindly: they form a design intent, modify
the original, and inspect whether the result matches that intent,
repeating until satisfied (Figure~\ref{fig:human_workflow}). Inspired by
this, ACE evaluates whether its edit \emph{from the original} followed the
instruction's intent---using an instruction-following (IF) score and a
natural-language reason (\S\ref{sec:metrics})---and re-iterates when the
score is below threshold (Figure~\ref{fig:overall}). Each iteration is a
single LLM round (\texttt{max\_turns}{=}1) that emits one batch of tool
calls and commits once, preventing duplicated destructive operations within
a turn. Canvas state \emph{accumulates} across iterations (later rounds
skip re-import), so corrections compound on the prior result. The IF
critique is forwarded \emph{verbatim}; we found it already actionable, and
distilling it into bullets did not help. The loop halts when IF reaches
$\tau{=}4$ or after $T{=}3$ iterations, and we use the final iteration's
result (Algorithm~\ref{alg:selfcorrect}). At deployment, a
\emph{strict-peak rollback} additionally returns an earlier iteration
whenever the critic's own logged score declines, removing the loop's
observed regressions with no ground truth required (\S\ref{sec:sc-exp}).

In Algorithm~\ref{alg:selfcorrect}, $s$ is the working scene-graph state,
$\tau$ the acceptance threshold and $T$ the maximum number of iterations,
and $(if_t, c_t)$ are the IF score ($if_t\in\{0,\dots,5\}$) and the
natural-language critique returned by the judge $\mathcal{J}$ at iteration
$t$. The judge never sees a ground-truth deck; it scores a symbolic edit
trace \textsc{JsonDiff}, computed by aligning the origin deck $D_0$ and the
current state $s$ via a stable per-node source id and labelling each
surviving / created / deleted node as a \emph{Modified} / \emph{Added} /
\emph{Removed} entry (with explicit slide-reorder and z-order entries for
id-paired elements that change position). The full computation---property
set, normalization, tolerances, and rendering for the judge---is given in
Appendix~\ref{app:jsondiff}, and the complete ACE design-agent and IF-judge
prompts are in Appendix~\ref{app:prompts}.

\begin{algorithm}[t]
\small
\caption{Self-correction with a GT-free IF judge}
\label{alg:selfcorrect}
\begin{algorithmic}[1]
\Require origin deck $D_0$, instruction $x$, agent $\mathcal{A}$, judge
$\mathcal{J}$, threshold $\tau{=}4$, max iters $T{=}3$
\State $s \gets \textsc{Import}(D_0)$
\For{$t = 1 \ldots T$}
  \State $s \gets \mathcal{A}(s, x;\ \texttt{max\_turns}{=}1)$
         \Comment{one round, batched ops, one commit}
  \State $(if_t, c_t) \gets \mathcal{J}\big(D_0,\ \text{(opt.)}\,\textsc{Render}(s),\ \textsc{JsonDiff},\ x\big)$
         \Comment{score + critique, no GT}
  \If{$if_t \ge \tau$} \State \textbf{break} \Comment{satisfied} \EndIf
  \State $x \gets \textsc{WrapCritique}(c_t)$;\ \ \texttt{skipImport}$\gets$\textbf{true}
\EndFor
\State \Return $s$ \Comment{final iteration}
\end{algorithmic}
\end{algorithm}

\section{Evaluation Suite and Protocol}
\label{sec:bench}

We introduce \textsc{figma-slide-bench-v1}, \textbf{97} human-authored
multi-slide editing tasks---each an original template, a target template,
and an instruction---adapted from PPTArena to the scene-graph setting;
\textbf{94} are automatically evaluable. The other three (Cases 102--104)
lack an origin deck (input is a template id, free text, or a reference
image) and require production-editor retrieval outside our deck-given scope,
so we exclude them from automatic evaluation and release them as future
work. For provenance, the suite \emph{extends} rather than replaces
PPTArena: from its 100 instructions we drop 15 (duplicates,
slide-unsupported, or overly broad; Appendix~\ref{app:data}) and re-curate
the remaining \textbf{85} (41 verbatim, 12 minor variants, 32 rewritten),
then add \textbf{12 novel} tasks (Cases 101--112)---9 editing tasks with no
PowerPoint analogue and the 3 generation/retrieval tasks above.

\subsection{Evaluation Metrics}
\label{sec:metrics}
Following PPTArena \citep{pptarena} we score each edit with two LLM judges.
\textbf{Instruction Following (IF).} The IF judge receives only the
structured data diffs (JSON/XML summaries) between slides, which forces it
to concentrate on content-level correctness rather than surface
aesthetics. Crucially, unlike PPTArena---which diffs \emph{ground-truth
vs.\ prediction} under a fixed target---our IF judge compares
\emph{original vs.\ prediction} against the instruction: because design
admits many valid outputs, we ask whether the user's intent was met, not
whether one reference was reproduced.
\textbf{Visual Quality (VQ).} The VQ judge receives only rendered
screenshots of the predicted (and reference) slides, with its context
engineered for aesthetics---alignment, layout, and style against a rubric.
For multi-slide edits we pass only the slides with salient changes, so the
judge concentrates on the edit rather than the full deck. We also remove
PPTArena's \texttt{style\_target} term, which penalizes outputs that
deviate from one prescribed style even when they are visually valid
(Figure~\ref{fig:qualitative}). The full judge prompts are in
Appendix~\ref{app:prompts}. Judge validity is assessed against a blind
human panel and out-of-loop judges in \S\ref{sec:judgevalid}.

\paragraph{Why our absolute scores differ from PPTArena's.}
PPTArena reports IF/VQ of 2.36/2.69 on its own data. Our numbers are not
directly comparable: we use a different judge (\texttt{gpt-5.5}), a
GT-free IF protocol, no \texttt{style\_target} term, and a re-rendered
Figma-Slides environment. We therefore run \emph{all} pipelines under one
identical protocol and report relative gaps.

\paragraph{Evaluation subsets.}
For a controlled head-to-head against the OpenXML baseline (which requires
PowerPoint), we use a \textbf{53-task subset} matched to PPTArena's
category distribution and converted to PPTX. By construction it contains
only adapted tasks and \emph{no} novel-only tasks, because the legacy
pipeline cannot represent them; we therefore report novel-task performance
on ACE separately (\S\ref{sec:agg}).

\section{Experiments}
\label{sec:exp}

\subsection{Setup}
\label{sec:setup}
We evaluate three backbones---\texttt{claude-sonnet-4-6}, \texttt{gpt-5.5},
and \texttt{gemini-3.5-flash}---and use \texttt{gpt-5.5} as the in-loop VLM judge; \S\ref{sec:judgevalid}
additionally re-scores identical outputs with two \emph{out-of-loop}
judges (\texttt{claude-sonnet-4-6}, \texttt{gemini-3.5-flash}) that play
no role in generation or the loop.
Most models are configured with \texttt{max\_tokens}$=$16{,}384; Gemini~3.5
Flash Preview uses \texttt{max\_tokens}$=$32{,}768 to accommodate its
extensive chain-of-thought reasoning. The agent is permitted up to
\textbf{3 iterations} and 35 agent turns per task, at \texttt{temperature}
0.0. We compare three pipelines: \textbf{ACE} (scene-graph edits,
design-engine render); \textbf{Claude-Skill HTML}, the Claude Code agent
equipped with a slide-editing Agent Skill that edits a per-deck
\texttt{index.html} rendered in Chromium using the \emph{identical}
backbone (\emph{the HTML pipeline} for short; construction in
Appendix~\ref{app:html}); and \textbf{PPTArena}
\citep{pptarena}, the OpenXML/\texttt{python-pptx} baseline rendered with
LibreOffice.

\subsection{Main Result}
\label{sec:main}
Table~\ref{tab:main} reports the 53-case head-to-head; agent-trajectory
statistics are in Appendix~\ref{app:traj}, and paired statistics with
CIs in \S\ref{sec:stats}. ACE leads on
\textbf{instruction following and on efficiency}, and its cost \emph{includes} the
self-correction loop: even so it is 1.75$\times$ faster and $\sim$44\%
cheaper than the agentic HTML pipeline (whose own iterative loop, large
HTML context, and browser renders dominate its cost; see
Appendix~\ref{app:cost} for the exact computation). With a \texttt{gpt-5.5}
backbone ACE is the strongest configuration overall and $\sim$7$\times$
cheaper than the HTML agent. ACE's VQ \emph{mean} advantage over HTML is
within statistical noise (\S\ref{sec:stats}), though blind raters prefer
ACE on VQ (57.1\%; \S\ref{sec:judgevalid}); the IF gap over PPTArena is
judge-invariant ($+2.1$--$2.4$, $p{<}0.001$;
Appendix~\ref{app:oojudge}). When a baseline produced no valid output
(HTML 3/53; PPTArena 20/53 unedited) we did \emph{not} assign a score of
0; instead the original render serves as the prediction, so its VQ
reflects the original-vs-reference similarity while its IF reflects the
unmet instruction (PPTArena's attempt rate and conditional quality are
decomposed in Appendix~\ref{app:pptarena}).

\begin{table}[t]
\centering\small
\adjustbox{max width=\columnwidth}{%
\begin{tabular}{lrrrr}
\toprule
                          & IF & VQ & Time(s) & Cost(\$) \\
\midrule
\textbf{ACE$_{\text{gpt}}$ (+SC)} & \textbf{4.74} & \textbf{4.19} & \textbf{112.3} & \textbf{0.134}\\
ACE (claude, +SC)        & 4.45 & 4.02 & 115.9 & 0.545\\
Claude-Skill HTML (agentic)     & 4.09 & 3.89 & 203.0 & 0.968\\
PPTArena (OOXML)               & 2.38 & 2.30 & \ \,80.4 & $\sim$0.04$^{\dagger}$\\
\bottomrule
\end{tabular}}
\caption{Four-way comparison on the 53-case subset.
\textbf{Backbone:} \texttt{claude-sonnet-4-6} for ACE and Claude-Skill
HTML;
\texttt{gpt-5.5} for ACE$_{\text{gpt}}$ and PPTArena.
\textbf{Judge:} \texttt{gpt-5.5}.
$^{\dagger}$Lower-bound estimate (per-iteration cost not logged). Scores
and cost are averaged over all 53 tasks per pipeline, with the do-nothing
fallback for no-output cases (\S\ref{sec:main};
Appendix~\ref{app:cost}).
\textbf{Coverage} (produced an edit): ACE 53/53, ACE$_{\text{gpt}}$
53/53, HTML 50/53, PPTArena 33/53 (attempt rate and conditional quality
decomposed in Appendix~\ref{app:pptarena}). All four pipelines are
agentic/multi-turn; only ACE's loop is guided by the IF judge
(circularity bounded in \S\ref{sec:judgevalid}).}
\label{tab:main}
\end{table}

\subsection{Judge Validity: Blind Human Study and Out-of-Loop Judges}
\label{sec:judgevalid}
The IF judge both guides self-correction and is a reported metric, so we
test it two ways on the \emph{identical, unchanged} outputs.

\paragraph{Blind human study.}
26 non-expert raters (after two pre-stated exclusions) cast \textbf{935}
blind win/tie/loss judgments on side-randomized pairs---51 ACE-vs-HTML
and 17 self-correction cases (protocol, exclusion rules, CIs, and
inter-rater agreement in Appendix~\ref{app:human}). The in-loop
\texttt{gpt-5.5} judge matches the blind human majority on decided cases
(ties excluded on both sides): \textbf{IF 80\%} ($n{=}41$), \textbf{VQ
76\%} ($n{=}37$), \textbf{Overall 78\%} ($n{=}50$), all $p{\le}.003$ vs.\ chance: it
tracks human perception, not a self-preference.
Humans independently reproduce both headline effects
(Table~\ref{tab:human}): win exceeds loss in every cell, decisively for
self-correction ($\approx$4:1; decisive win-rate 81\%, $p{<}0.001$,
\emph{including} on VQ, which never gates the loop) and significantly for
ACE-vs-HTML at the same backbone (IF 59.6\% / VQ 57.1\% / Overall
58.7\% decisive win-rates, all $p{\le}.0025$).

\begin{table}[t]
\centering\small
\adjustbox{max width=\columnwidth}{%
\begin{tabular}{@{}lccc@{}}
\toprule
Human preference (W/T/L \%) & IF & VQ & Overall \\
\midrule
ACE vs.\ Claude-Skill HTML      & 31/48/21 & 37/35/28 & 41/31/29 \\
Self-corrected vs.\ single-pass & 47/42/11 & 44/47/9  & 48/41/11 \\
\bottomrule
\end{tabular}}
\caption{Blind pairwise human preference (26 raters, 935 judgments); win
exceeds loss in every cell. Win-rates, CIs, protocol:
Appendix~\ref{app:human}.}
\label{tab:human}
\end{table}

\paragraph{Out-of-loop judges.}
Re-scoring the identical outputs with \texttt{claude-sonnet-4-6} and
\texttt{gemini-3.5-flash}---neither touches generation or the loop---leaves
the ranking ACE $>$ Claude-Skill HTML $>$ PPTArena invariant across
judges and
backbones (Table~\ref{tab:oojudge}). The self-correction
gain likewise survives: $\Delta$IF $+0.94$ (in-loop) $\to$ $+0.61$
(claude) $\to$ $+0.56$ (gemini), with VQ rising too
($+0.78/{+}0.56/{+}1.33$); a 35-task identical-output control bounds
cross-run judge noise at $\le{+}0.17$ IF, far below the gain
(Appendix~\ref{app:oojudge}). Out-of-loop judges thus recover roughly
two-thirds of the in-loop gain, bounding any critic-specific component at
about one-third of the measured effect.

\begin{table}[t]
\centering\small
\adjustbox{max width=\columnwidth}{%
\begin{tabular}{@{}lccc@{}}
\toprule
 & \multicolumn{3}{c}{IF\,/\,VQ by judge} \\
\cmidrule(l){2-4}
System (backbone) & gpt-5.5 (in-loop) & claude-4.6 & gemini-3.5 \\
\midrule
ACE (gpt-5.5)           & 4.74\,/\,4.19 & 4.55\,/\,3.70 & 4.79\,/\,4.21 \\
ACE (claude-4.6)        & 4.45\,/\,4.02 & 4.23\,/\,3.49 & 4.77\,/\,4.25 \\
Claude-Skill HTML (claude-4.6) & 4.09\,/\,3.89 & 4.09\,/\,3.34 & 4.45\,/\,4.02 \\
PPTArena (gpt-5.5)      & 2.38\,/\,2.30 & 2.42\,/\,2.25 & 2.40\,/\,2.28 \\
\bottomrule
\end{tabular}}
\caption{Identical outputs re-scored by two out-of-loop judge families
($n{=}53$): the ranking ACE $>$ Claude-Skill HTML $>$ PPTArena is
judge-invariant.}
\label{tab:oojudge}
\end{table}

\subsection{Paired Statistics on the Full Benchmark}
\label{sec:stats}
The 53-case set is a subset; the complete \emph{edit} benchmark is 94
tasks (97 minus three generation-from-scratch cases). Claude-Skill HTML
predictions for the remaining 41 were generated and judged under the same
harness, judge, and protocol, so 94 completes coverage rather than
searching for significance. In Table~\ref{tab:stats}, the IF advantage is
significant on the full benchmark under both judges (gpt-5.5 at
$n{=}94$: also paired-$t$ $p{=}.009$, sign test $p{=}.04$, W/T/L
35/38/21) and
non-significant on the 53-subset under either---an underpowering artifact
(same sign and magnitude), not an absent effect.
VQ mean scores are statistically indistinguishable in every cell (all CIs
include 0); we state VQ exactly that way, noting that the blind panel
(\S\ref{sec:judgevalid}) resolves a consistent human VQ preference
(57.1\%) that a coarse 5-point mean cannot.

\begin{table}[t]
\centering\small
\adjustbox{max width=\columnwidth}{%
\begin{tabular}{@{}llrrrrlr@{}}
\toprule
Judge & M & Set & ACE & HTML & $\Delta$ & 95\% CI & $p$ \\
\midrule
gpt-5.5    & IF & 53 & 4.45 & 4.09 & $+$0.36 & [$-$0.06, $+$0.81] & .20 \\
gpt-5.5    & IF & 94 & 4.23 & 3.81 & $+$0.43 & [$+$0.13, $+$0.75] & \textbf{.010} \\
gemini-3.5 & IF & 53 & 4.77 & 4.45 & $+$0.32 & [$-$0.08, $+$0.77] & .15 \\
gemini-3.5 & IF & 94 & 4.62 & 4.21 & $+$0.40 & [$+$0.07, $+$0.76] & \textbf{.022} \\
gpt-5.5    & VQ & 94 & 3.66 & 3.57 & $+$0.09 & [$-$0.23, $+$0.40] & .56 \\
gemini-3.5 & VQ & 94 & 3.93 & 3.78 & $+$0.15 & [$-$0.27, $+$0.56] & .57 \\
\bottomrule
\end{tabular}}
\caption{ACE vs.\ Claude-Skill HTML, same \texttt{claude-sonnet-4-6}
backbone:
paired bootstrap 95\% CIs, two-sided Wilcoxon $p$. IF is significant at
$n{=}94$ under both judges; the 53-subset is underpowered; VQ means are
indistinguishable throughout.}
\label{tab:stats}
\end{table}

\subsection{Effect of Self-Correction}
\label{sec:sc-exp}
Self-correction is the main driver of ACE's quality. Without it
(Table~\ref{tab:abl}), ACE on \texttt{claude-sonnet-4-6} scores 4.04/3.75
in a single turn---already nearly matching the agentic HTML pipeline's
4.09/3.89. The loop then adds \textbf{+0.41 IF} and \textbf{+0.27 VQ}
whole-benchmark, lifting ACE to 4.45/4.02. The average understates the
mechanism, because correction is selective: 35/53 tasks (66\%) reach the
threshold at iteration~1 (mean IF 4.60) and never enter the loop;
conditioned on the 18 that do enter, 13 (72\%) improve---by \textbf{+0.94 IF / +0.78 VQ} on
average---2 are unchanged, and 3 regress by $\le$1 point each. A
\emph{strict-peak rollback}---return the earlier iteration whenever the
critic's own logged score declined---uses only the in-loop critic (no
ground truth) and removes every regression: IF 4.45$\to$4.49, VQ 4.02$\to$4.06, with no task harmed
(Appendix~\ref{app:sens}). Because VQ is \emph{not} part of the stopping
signal, its rise---preserved under out-of-loop judges
($+0.56/{+}1.33$; Appendix~\ref{app:oojudge}) and confirmed by an
82\% blind human preference (\S\ref{sec:judgevalid})---is independent
evidence that corrections improve genuine design quality, not only the
optimized IF metric. The gain is also insensitive to the loop's two
knobs: replayed curves over the iteration budget $K$ and halt threshold
$\tau$ are monotone with plateaus ($\sim$65\% of the benefit by $K{=}2$;
Appendix~\ref{app:sens}).

\subsection{Ablations and Component Isolation}
\label{sec:abl}
Table~\ref{tab:abl} isolates the representation with self-correction
\emph{disabled}. Replacing the scene-graph with OpenXML is uniformly and
substantially worse across all three backbones, confirming that the
\emph{representation}, not merely the toolset, carries the gains.

Beyond the representation, a leave-one-out isolation (Table~\ref{tab:iso},
Appendix~\ref{app:careaudit}) removes each remaining component with the
other four and the evaluation held fixed, scored where the component is
active. Removing content-aware routing costs up to $-0.75$ IF; removing the
specialized action space costs $-0.55$ to $-1.00$ IF on the tasks that
invoke those tools, with operation counts inflating $\approx$1.8$\times$
as charts and tables are rebuilt from primitives; and removing
self-correction costs $-0.41$ IF---each component contributes
independently. CARE, in particular, is not only a cost lever:
on the 16 multi-slide tasks where routing reduces context, substituting
the full deck lowers IF on two of three backbones ($-0.75$ claude /
$-0.38$ gpt; gemini flat), since a focused context keeps the model on the
target edit. A direct routing audit finds \textbf{52/53}
correct mode decisions (1 under-scope, 0 over-scope), perfect
slide-selection recall (20/20), and 13/13 on explicitly named slides; the
single under-scope (Case~75) is recovered by self-correction (IF
3.0$\to$4.0). Audit method, per-case labels, overflow and deck-size
controls, and the isolation construction are in
Appendix~\ref{app:careaudit}.

Crucially, the scene-graph also wins on \emph{cost}, not only quality: under
identical CARE routing, the scene-graph representation uses \textbf{15--31\%
fewer input tokens per call} than the OpenXML one (12--67\% per case, as the
saving compounds with fewer agent turns), directly lowering API cost;
Appendix~\ref{appx:input_tokens} reports the per-model breakdown.

\begin{table}[t]
\centering\small
\adjustbox{max width=\columnwidth}{%
\begin{tabular}{l cc}
\toprule
& \textbf{ACE} & \textbf{OpenXML} \\
\textbf{Backbone} & (SG) & (XML) \\
& IF\,/\,VQ & IF\,/\,VQ \\
\midrule
\texttt{claude-sonnet-4-6} & 4.04\,/\,3.75 & 3.40\,/\,3.13\\
\texttt{gpt-5.5}           & 4.60\,/\,4.17 & 3.91\,/\,3.72\\
\texttt{gemini-3.5-flash}  & 4.06\,/\,3.91 & 3.28\,/\,3.28\\
\bottomrule
\end{tabular}}
\caption{Ablation over representation (53 cases; mean IF\,/\,VQ),
\textbf{self-correction disabled}. SG = scene-graph. Replacing the
scene-graph with OpenXML is uniformly worse. The ACE (SG) column is the
single-pass result; self-correction raises \texttt{claude-sonnet-4-6} to
4.45/4.02 (Table~\ref{tab:main}).}
\label{tab:abl}
\end{table}

\subsection{Qualitative Results}
\label{sec:qual}

\begin{figure}[t]
  \centering
  \setlength{\tabcolsep}{1.5pt}
  \renewcommand{\arraystretch}{1.0}
  \newcommand{\mcell}[1]{\includegraphics[width=0.235\columnwidth,height=1.7cm,keepaspectratio]{latex/images/#1}}
  \begin{tabular}{@{}cccc@{}}
    \scriptsize Reference & \scriptsize ACE (ours)
      & \scriptsize Claude-Skill & \scriptsize PPTArena \\[1pt]
    \mcell{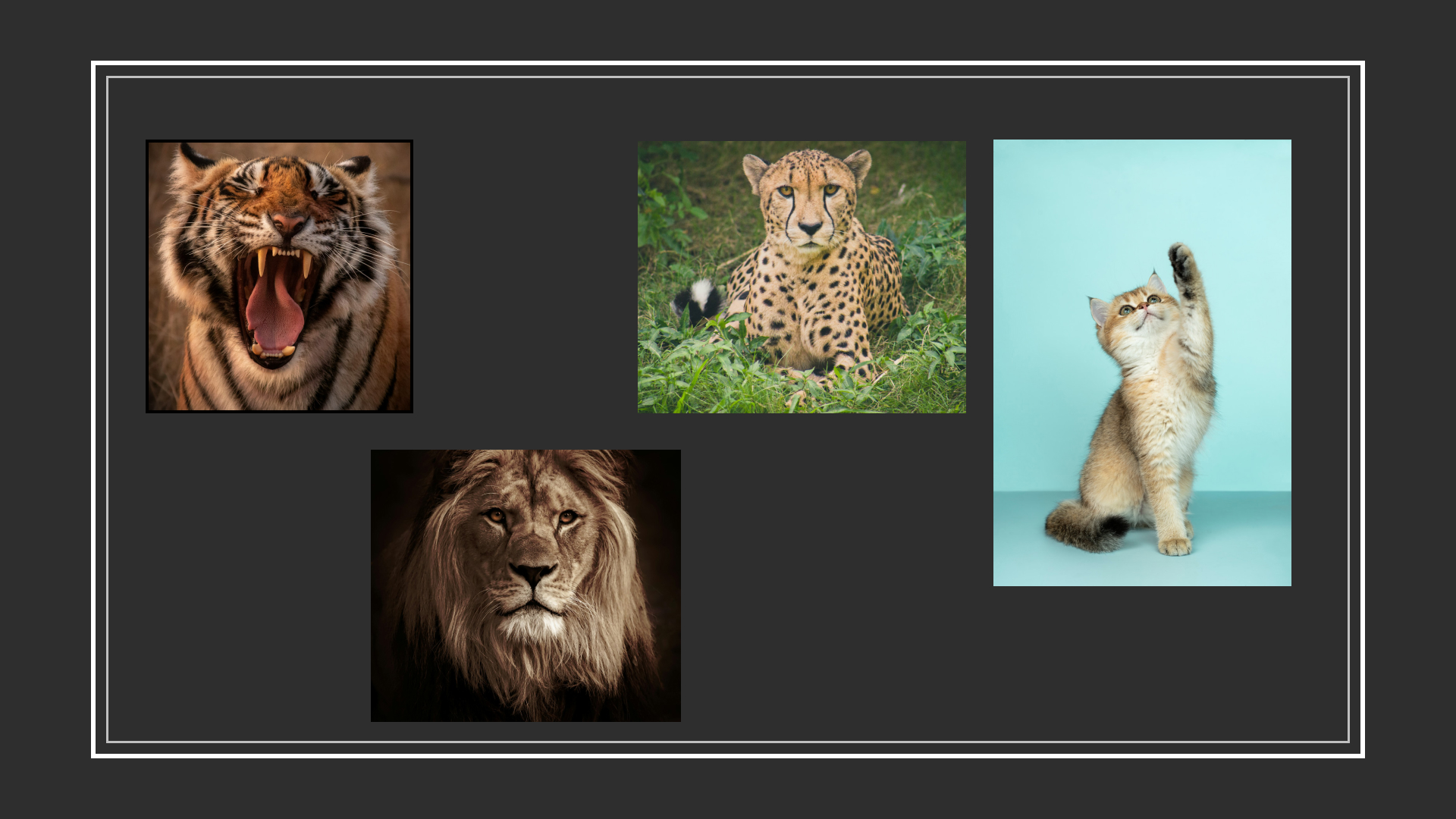} & \mcell{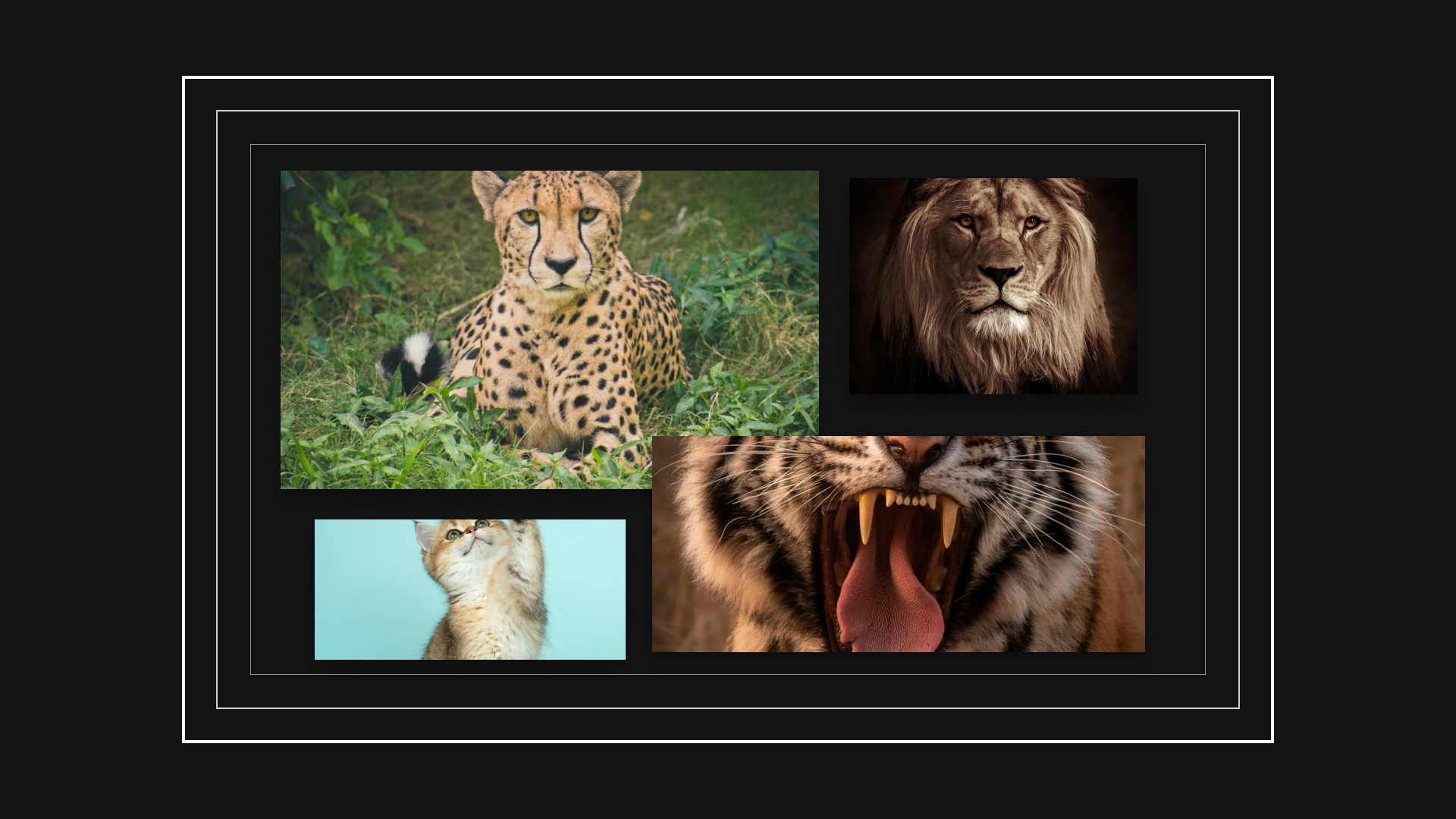}
      & \mcell{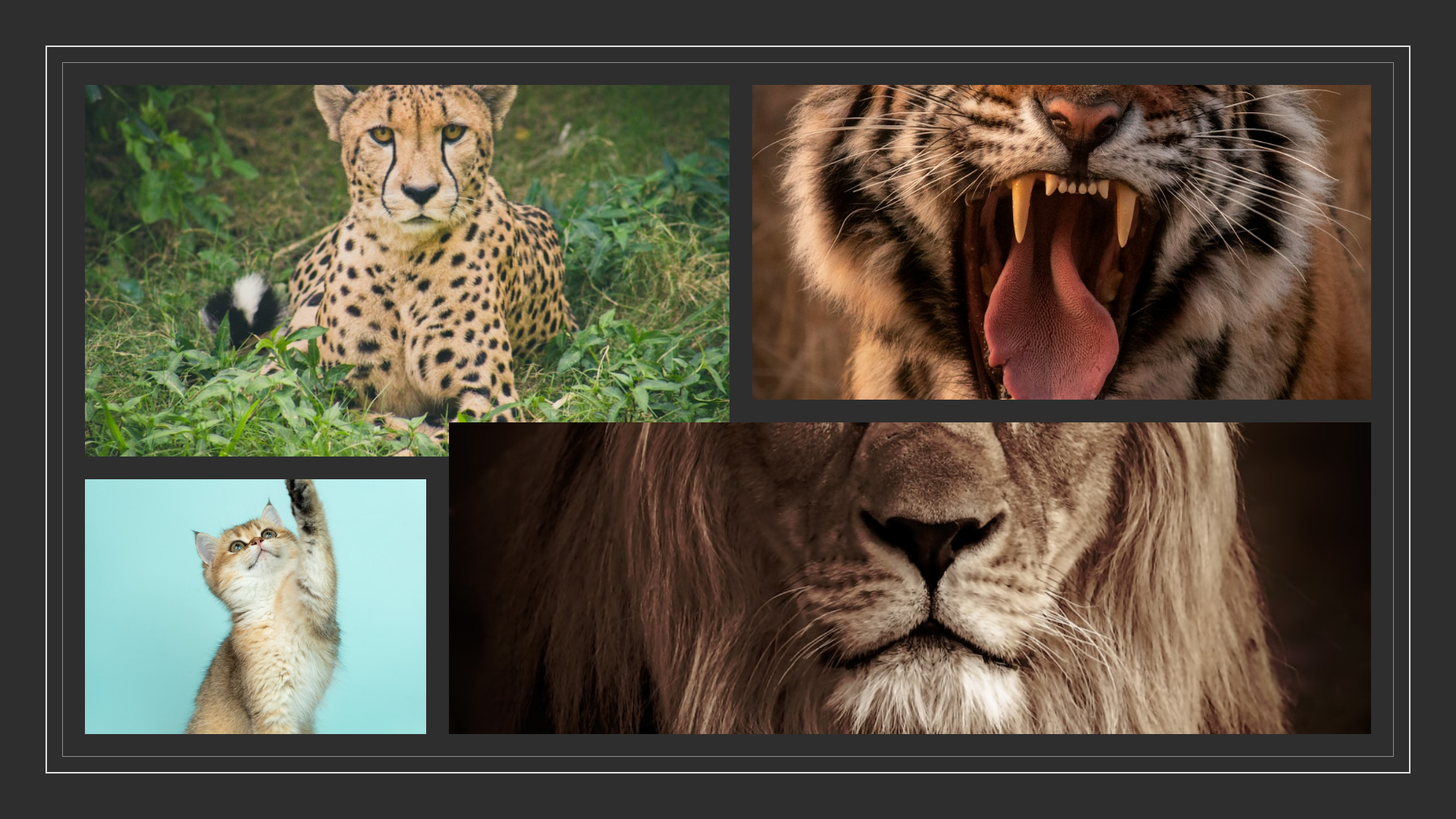} & \mcell{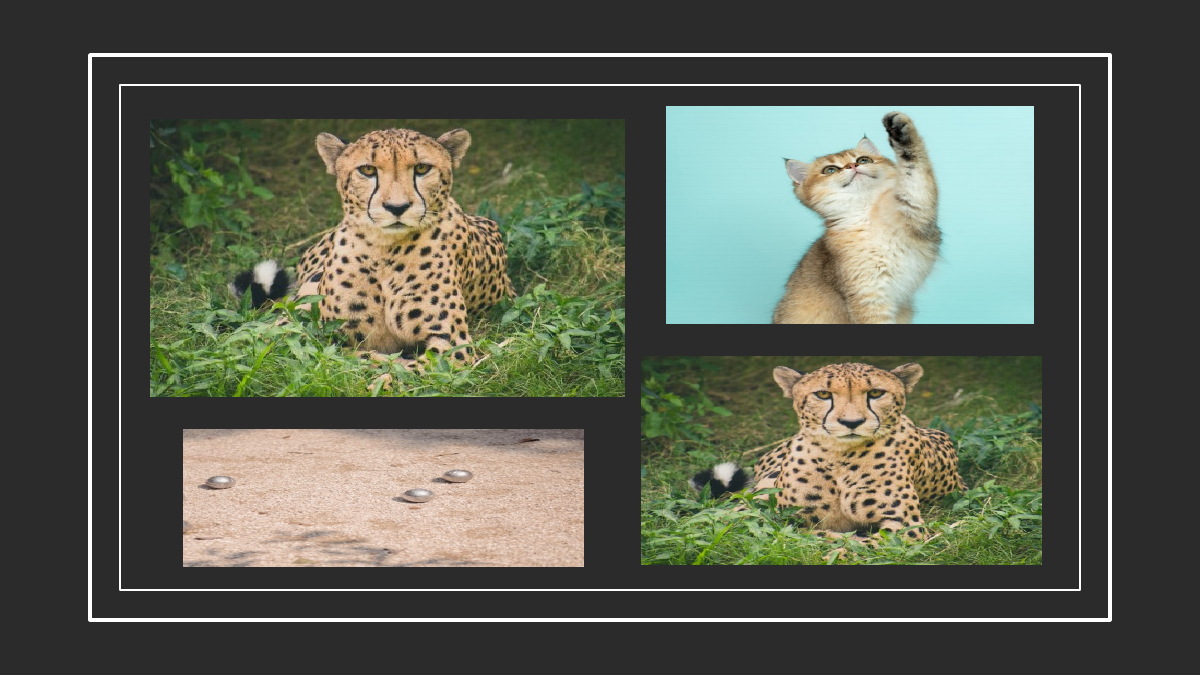} \\
  \end{tabular}
  \caption{Qualitative example (Case~17, build ensemble category boards).
    ACE's edit differs from the manual reference yet remains valid---what
    our reference-free judge rewards---while PPTArena fails to build the
    grid. Backbones: \texttt{claude-sonnet-4-6} (ACE/HTML);
    \texttt{gpt-5.5} judge. Further comparisons: Appendix~\ref{app:qual}.}
  \label{fig:qualitative}
\end{figure}

Because design has no unique ground truth, ACE often produces edits that
\emph{differ} from the reference yet still satisfy the instruction---what
our reference-free judge rewards. On Case~17 (Figure~\ref{fig:qualitative}),
ACE and HTML both produce valid boards unlike the reference while PPTArena
fails; the more discriminative Case~71---ACE colours exactly the negative
cells, HTML over-applies to whole rows, and PPTArena leaves the table
unchanged---and four further tasks appear in Figure~\ref{fig:qual_appendix}
on page~\pageref{fig:qual_appendix} (further discussion in
Appendix~\ref{app:qual}).

\subsection{Aggregate Results}
\label{sec:agg}
On the 9 novel editing tasks ACE (with self-correction) attains IF 3.78 /
VQ 3.22---below its 4.45/4.02 on the adapted subset, so the novel tasks
are unsaturated---and IF 4.23 / VQ 3.66 on the full 94-task benchmark
(paired CIs vs.\ HTML in \S\ref{sec:stats}); per-task novel and
per-category breakdowns are in Appendix~\ref{app:fullresults}.

\label{endofmain}
\section*{Limitations}
\textbf{Comparison fairness.} All pipelines are agentic and
multi-turn---the HTML pipeline and the OpenXML baseline both iterate
internally---so iteration count is not the confound, and the leave-one-out
ablations (\S\ref{sec:abl}) now attribute the gains component-by-component
at a fixed backbone and judge. The remaining asymmetry is the \emph{signal}
guiding each loop: only ACE's loop is guided by the same judge family used
for one metric, which is the circularity we bound below.
\textbf{Judge circularity---quantified, not eliminated.} The IF judge both
drives self-correction and is the IF metric. The blind human study and the
out-of-loop judge re-scoring (\S\ref{sec:judgevalid}) independently
support the reported gains and bound any critic-specific component at
roughly one-third of the measured effect. The in-loop critic nonetheless
shares a model family with one reported metric, and we retain VQ (never a
stopping signal) and the released qualitative comparisons
(Appendix~\ref{app:qual}) as further independent checks.
\textbf{VQ.} Against the HTML pipeline, VQ \emph{mean} differences are
statistically indistinguishable (\S\ref{sec:stats}); we report them
exactly as such, while the blind panel shows a modest but consistent human
VQ preference for ACE. Interactivity remains the one weak category
(Table~\ref{tab:percat}).
\textbf{Benchmark provenance.} The suite is largely adapted from PPTArena
(85/97 tasks, 41 verbatim); 12 are novel, of which 9 are novel editing
tasks with no PowerPoint analogue. The head-to-head subset contains no
novel-only tasks, so novel-task results are ACE-only.
\textbf{Coverage.} Three tasks (102--104) are not automatically evaluable:
they have no origin deck and require production-side template retrieval
(by id, keyword, or style), which is outside our deck-given scope and left
to future work.
\textbf{Platform transfer is engineering future work.} Nothing in the
method is Figma-exclusive: a hierarchical shape/element tree is exactly
how PowerPoint's OOXML shape tree and the Google Slides
\texttt{pageElements} API model a slide, and our OOXML ablation
(Table~\ref{tab:abl}) is precisely the flat, PowerPoint-like
condition---ACE already runs on it, at a quantified 0.6--0.8 IF cost that
measures what auto-layout is worth. CARE and the critic are
substrate-agnostic: they need only a serializable deck representation and
a renderer, both of which PowerPoint (\texttt{python-pptx}/LibreOffice)
and Google Slides (API export) provide---this paper already renders OOXML
via LibreOffice and HTML via Chromium. Re-binding the 98 tools to each
platform's API and mapping auto-layout onto placeholder layouts is
engineering we scope explicitly as future work.
\textbf{Subjective prompts} without explicit design-system constraints
yield high output variance.

\section*{Ethics Statement}
All human faces in the benchmark assets were replaced with generic avatars
or royalty-free stock photos, and design assets are used under CC-BY-4.0.
The blind human study (\S\ref{sec:judgevalid}) used adult volunteer raters
who evaluated anonymized slide renders; no personal data was collected.
The system is intended to assist, not replace, human designers.

\section*{Acknowledgments}
The researchers at Seoul National University (SNU) were supported by
grants from the Institute of Information \& Communications Technology
Planning \& Evaluation (IITP), funded by the Korean government, under
Grant Nos.\ RS-2021-II211343 and RS-2025-25442338.


\bibliography{custom}

\appendix

\section{Action-Space Modules}
\label{app:tools}
The 98-tool suite is registered across 11 modules
(Table~\ref{tab:modules}), extending the 52-tool Canvas UI action space
\citep{canvas} with presentation-specific operations.

\begin{table}[h]
\centering\small
\adjustbox{max width=\columnwidth}{%
\begin{tabular}{l r}
\toprule
\textbf{Module} & \textbf{Tools} \\
\midrule
\texttt{contentTools}      & 56 \\
\texttt{slideTools}        & 13 \\
\texttt{chartTools}        & \ 7 \\
\texttt{dataChartTools}    & \ 4 \\
\texttt{smartArtTools}     & \ 4 \\
\texttt{tableTools}        & \ 4 \\
\texttt{connectionTools}   & \ 3 \\
\texttt{importExportTools} & \ 3 \\
\texttt{unsplashTools}     & \ 2 \\
\texttt{batchTools}        & \ 1 \\
\texttt{mathTools}         & \ 1 \\
\midrule
\textbf{Total}             & \textbf{98} \\
\bottomrule
\end{tabular}}
\caption{Action-space modules. \texttt{importExportTools} includes
\texttt{clear\_canvas}, \texttt{export\_json}, \texttt{import\_json};
\texttt{batchTools} is \texttt{batch\_execute}; \texttt{mathTools} is
\texttt{create\_math}.}
\label{tab:modules}
\end{table}

\paragraph{Utilization audit.}
Expanding \texttt{batch\_execute} into its nested primitive commands
across all logged trajectories and backbones, ACE invokes \textbf{66 of
the 98 tools (67\%)} on the benchmark (per-backbone 46--57). The 32 unused
tools are not missing capabilities but tools the task mix never needs: 10
non-edit utilities and read-only getters; 9 post-hoc mutators and subtype
alternates made redundant by the one-shot \texttt{create\_*} constructors
that \emph{were} used; and 13 low-frequency shape/style primitives no
benchmark task demands. The suite is task-gated, not padded. Sub-element
control is exercised in real trajectories: sub-string text styling
(\texttt{set\_text\_range\_style}, \texttt{set\_text\_decoration}),
per-child responsive-layout properties (\texttt{set\_padding},
\texttt{set\_item\_spacing}, \texttt{set\_axis\_align}), and individual
visual setters (\texttt{replace\_color}, \texttt{set\_corner\_radius},
\texttt{set\_drop\_shadow}, \texttt{rotate\_node}).

\section{CARE Router: Control Flow}
\label{app:router}
The router consumes a \texttt{PresentationSummary} (\texttt{slide\_count}
and \texttt{slide\_names}, parsed from \texttt{meta.json}) and produces a
\texttt{RoutingDecision}. It first computes a deterministic heuristic
baseline $H$ (which also fixes the \texttt{needs\_image} flag), then runs a
single lightweight LLM call that classifies the routing \emph{mode} and
target slides. The LLM mode/targets are adopted only if they parse and
validate; otherwise they fall back to $H$. A final post-rule reclassifies
single-slide decks before the decision reaches \textsc{prepare\_context}
(Algorithm~\ref{alg:router}; full prompt in
Listing~\ref{lst:mode-prompt}).

\begin{algorithm}[h]
\small
\caption{\textsc{classify\_instruction\_single\_llm}}
\label{alg:router}
\begin{algorithmic}[1]
\Require instruction, summary, key, model
\If{LLM unavailable \textbf{or} key missing}
   \State \Return \textsc{heuristic}(instruction, summary)
\EndIf
\State $H \gets \textsc{heuristic}(\text{instruction, summary})$
\State $f_M \gets \textsc{submit}(\textsc{ModeLLM})$
\State $(\text{mode, targets}) \gets (H.\text{mode}, H.\text{targets})$
\State \textbf{try} $r \gets f_M.\text{result}()$; \textbf{if} valid \textbf{then} update mode/targets
\State $\text{needsImage} \gets H.\text{needsImage}$
\State \Return \textsc{maybeReclassifySingleSlide}(mode, targets, needsImage)
\end{algorithmic}
\end{algorithm}

\subsection{Heuristic Baseline $H$}
\label{app:router-heuristic}
\textsc{classify\_instruction} is a deterministic, LLM-free classifier that
always emits a complete \texttt{RoutingDecision}; it therefore acts as the
per-field fallback for the LLM output, and it is also the sole source of
the \texttt{needs\_image} flag (no separate visual-gating LLM is used). It
runs six banks of case-insensitive regular expressions over the lowercased
instruction: (i) \emph{global-scope} cues (``entire/whole/all/every
presentation''), (ii) \emph{systemic-token} cues (``color scheme'',
``theme'', ``palette'', ``dark/light mode''), (iii) \emph{macro/batch} cues
(``translate'', ``proofread'', ``find and replace'', ``bold all''),
(iv) \emph{structural} cues (``wrong order'', ``reorder/merge slides''),
(v) \emph{micro-spatial} cues (``add a chart'', ``move/resize/align'',
``z-order''), and (vi) two image-gating banks that set \texttt{needs\_image}
(an image-needed bank and a no-image bank, defaulting to \texttt{true} on
ties since missing visual context is more harmful than extra tokens). A
separate extractor resolves explicit slide references into 0-based indices
clamped to $[0,\texttt{slide\_count})$ (code reference). $H$ then selects a
mode in fixed priority order with an attached confidence:
systemic$\,\wedge\,$global $\rightarrow$ \textsc{Systemic\_Token}; any
structural cue $\rightarrow$ \textsc{Macro\_Programmatic} (whole deck);
macro$\,\wedge\,(>3$ targets$)\rightarrow$ \textsc{Macro\_Programmatic};
$1$--$6$ explicit targets $\rightarrow$ \textsc{Micro\_Spatial}; ambiguous
global $\rightarrow$ \textsc{Macro\_Programmatic}; no match $\rightarrow$
\textsc{Full}.

\subsection{Sampling Temperature}
\label{app:router-temp}
$T$ is the decoding (softmax) temperature of the routing LLM. We set
$T{=}0$ (greedy decoding), so routing is deterministic and reproducible and
adds no variance to downstream context selection.

\subsection{Mode Classification}
\label{app:router-mode}
The mode LLM maps the instruction to one of the three modes and extracts the
referenced slide indices, driven by \texttt{\_MODE\_LLM\_SYSTEM\_PROMPT}
(Listing~\ref{lst:mode-prompt}). The prompt defines each mode, enumerates
cross-slide-dependency cases that must \emph{not} be \textsc{Micro\_Spatial}
(e.g.\ ``match the deck's style''), and requires source/reference slides to
be included in \texttt{target\_slides}. The model returns a single JSON
object \{\texttt{mode}, \texttt{target\_slides}, \texttt{reason}\}; it is
adopted only if it parses and \texttt{mode} is valid, with indices clamped
to $[1,\texttt{slide\_count}]$ and shifted to 0-based, else the field falls
back to $H$.

\subsection{Single-Slide Reclassification}
\label{app:router-single}
\textsc{maybe\_reclassify\_single\_slide} guards the degenerate case: if the
deck has exactly one slide and the mode is \textsc{Macro\_Programmatic}, it
is rewritten to \textsc{Micro\_Spatial} with \texttt{target\_slides}${=}[0]$
(preserving \texttt{reason}, confidence, and \texttt{needs\_image}), since a
whole-deck batch operation collapses to a single spatial edit on a one-slide
deck. The corrected decision is handed to \textsc{prepare\_context}.

\begin{lstlisting}[style=prompt,caption={\texttt{\_MODE\_LLM\_SYSTEM\_PROMPT}: mode-classification system prompt.},label={lst:mode-prompt},float=*]
You are a routing engine for a Figma Slides editing system.
Given a user instruction and a presentation summary, classify the instruction into EXACTLY ONE mode and extract target slides.

## Mode Classification

MICRO_SPATIAL
  - Targets specific slide(s) by number/position AND all needed info is within those slides.
  - Operations: add/edit/move/resize elements, spatial layout, single-slide chart/table creation.
  - Do NOT use MICRO_SPATIAL if the instruction needs context from OTHER, non-targeted slides:
    * "match the presentation's style/colors/theme" -> SYSTEMIC_TOKEN
    * "make it consistent with the rest of the deck" -> SYSTEMIC_TOKEN or MACRO_PROGRAMMATIC
    * "use the same format as slide 1 on slide 5" -> include both in target_slides
  - Example: "On slide 5, add a bar chart"; "Move the title on the last slide".

MACRO_PROGRAMMATIC
  - Applies an operation across many/all slides, OR whole-presentation structural changes.
  - Operations: translate all text, find-and-replace, change font everywhere, proofread,
    bold all titles, delete/merge/consolidate slides, reorder slides, fix slide order.
  - Any slide reordering / fixing order / aligning an agenda with slides is ALWAYS MACRO_PROGRAMMATIC.
  - Example: "Translate the entire presentation"; "The slides are in the wrong order - fix it".

SYSTEMIC_TOKEN
  - Changes design-system-level properties (colors, themes, palettes) globally.
  - Example: "Apply a dark mode theme"; "Change the accent color to blue across all slides".

## Target Slides (1-based)
Extract ALL slide numbers mentioned, including BOTH modified slides AND source/reference slides.
- "Change all headings. On slides 3 and 8, resize the image." -> [3, 8]
- "Translate the entire presentation." -> []
- "Apply the layout from slide 1 to slides 2 through 5." -> [1, 2, 3, 4, 5]  (slide 1 is the source!)
- If a slide is a source/template/example ("from slide 1", "like slide 3"), you MUST include it.

Respond with ONLY a JSON object:
{"mode": "MICRO_SPATIAL" | "MACRO_PROGRAMMATIC" | "SYSTEMIC_TOKEN", "target_slides": [1, 5, 9], "reason": "<one sentence>"}
\end{lstlisting}

\section{Per-Call Input-Token Reduction: Scene-Graph vs.\ OpenXML}
\label{appx:input_tokens}

\begin{table}[h]
\centering\small
\adjustbox{max width=\columnwidth}{%
\begin{tabular}{lrrr}
\toprule
Model & XML in/call & Ours in/call & Reduction \\
\midrule
\texttt{claude-sonnet-4-6} & 62.4K & 48.1K & 22.9\% \\
\texttt{gpt-5.5}           & 38.4K & 32.7K & 14.9\% \\
\texttt{gemini-3.5-flash}  & 55.3K & 38.4K & 30.6\% \\
\bottomrule
\end{tabular}}
\caption{Per-call input tokens for ACE on the scene-graph (\emph{Ours})
vs.\ the identical agent and CARE routing on an OpenXML/XML serialization
(53 cases, iter-1).}
\label{tab:percall}
\end{table}

\noindent\textbf{Per-call (15--31\%):} for the same CARE-routed content,
the scene-graph serialization is more compact than OpenXML---largest
reduction on \texttt{micro\_spatial} edits, smallest on
\texttt{systemic\_token}.
\textbf{Per-case (12--67\%):} per-call savings compound with fewer turns
(e.g.\ \texttt{gemini} 4.8$\to$3.0 calls).
\textbf{The \texttt{gpt-5.5} gap is smallest} because its XML baseline is
already lean (38K vs.\ claude's 62K), leaving less to slim down.

\section{Cost and Timing Computation}
\label{app:cost}
Table~\ref{tab:main} reports per-case wall-clock time and cost. All four
pipelines are agentic and multi-turn, so we count \emph{every} model call:
agent turns for all systems, plus---for \textbf{ACE}---the in-loop IF judge
and the one-shot CARE router. Failed/no-output cases are filled by the
do-nothing baseline (original render as prediction; \S\ref{sec:main}); time
and cost are averaged over each pipeline's logged cases.
Tables~\ref{tab:cost_gpt} and~\ref{tab:cost_claude} break down the per-case
ACE agent cost from provider usage records (uncached input, cache read,
cache write, output) at list rates; recomputing from unit prices reproduces
the Table~\ref{tab:main} totals (\$0.134 and \$0.545).

\begin{table}[h]
\centering\small
\adjustbox{max width=\columnwidth}{%
\begin{tabular}{lrrr}
\toprule
Token type & tokens/case & \$/1M & \$/case \\
\midrule
input (uncached) & 38{,}397  & 1.25  & 0.0480 \\
cache read       & 168{,}873 & 0.125 & 0.0211 \\
output           & 4{,}114   & 10.00 & 0.0411 \\
\midrule
Agent subtotal   &           &       & 0.1102 \\
\quad + judge + router &     &       & 0.0247 \\
\textbf{Total}   &           &       & \textbf{0.134} \\
\bottomrule
\end{tabular}}
\caption{Per-case ACE cost, \texttt{gpt-5.5} backbone (53-case average).}
\label{tab:cost_gpt}
\end{table}

\begin{table}[h]
\centering\small
\adjustbox{max width=\columnwidth}{%
\begin{tabular}{lrrr}
\toprule
Token type & tokens/case & \$/1M & \$/case \\
\midrule
input (uncached)  & 49{,}312  & 3.00  & 0.1479 \\
cache read        & 378{,}356 & 0.30  & 0.1135 \\
cache write (5m)  & 49{,}158  & 3.75  & 0.1843 \\
output            & 4{,}830   & 15.00 & 0.0725 \\
\midrule
Agent subtotal    &           &       & 0.5182 \\
\quad + judge + router &      &       & 0.0270 \\
\textbf{Total}    &           &       & \textbf{0.545} \\
\bottomrule
\end{tabular}}
\caption{Per-case ACE cost, \texttt{claude-sonnet-4-6} backbone (53-case average).}
\label{tab:cost_claude}
\end{table}

\paragraph{Why \texttt{gpt-5.5} is cheaper.} For \texttt{gpt-5.5}, $81\%$ of
input tokens are cache reads (billed at $1/10$ the input rate) and its unit
prices are $\sim$2.4$\times$ lower than \texttt{claude-sonnet-4-6}'s; OpenAI
also charges \emph{nothing} for cache writes. For \texttt{claude-sonnet-4-6}
the single largest line is, perhaps surprisingly, cache \emph{write}
(\$0.184)---Anthropic prices cache creation at $1.25\times$ the input
rate---which, together with uncached input (\$0.148) and cache reads
(\$0.114), drives the \$0.518 agent cost.

\paragraph{Baselines.} The Claude-Skill HTML averages (\$0.968, $203.0$\,s over the
$50$ logged cases) come from each run's logged total, which already includes
its iterative loop, large HTML context, and browser renders. PPTArena's cost
($\sim$\$0.04, averaged over the 27 cases with usage logs; the audited
edit count is 33/53, Appendix~\ref{app:pptarena}) is a deliberate
\emph{lower bound}: per-iteration usage is not logged, so we price a
single forward pass at \texttt{gpt-5} rates, and the true cost is
plausibly $1.5$--$3\times$ higher. Counting the full self-correction loop,
\textbf{ACE}\textsubscript{gpt} (\$0.134) is thus $\sim$7$\times$ cheaper
than the HTML agent while leading on instruction following under all three
judges (Table~\ref{tab:oojudge}).

\section{The \textsc{JsonDiff} Edit Trace}
\label{app:jsondiff}

Algorithm~\ref{alg:selfcorrect} feeds the judge $\mathcal{J}$ a symbolic
representation of the agent's edits, denoted \textsc{JsonDiff}. This
appendix specifies how it is computed and rendered.

\subsection{Origin-vs-Current, not GT-vs-Prediction}
\textsc{JsonDiff} compares the imported origin deck $D_0$ against the
agent's current document state $s$, and \emph{never} against a ground-truth
deck. The plugin that imports and exports Figma documents preserves a
stable source id on every slide and node end-to-end
(\texttt{setPluginData(`sourceId')}), so $D_0$ and $s$ live in a single id
namespace: a node that survived an edit keeps its id on both sides, a
created node carries a fresh id present only in $s$, and a deleted node
appears only in $D_0$. The diff therefore reads exactly as ``what the agent
changed.'' Each entry is labelled from the agent's perspective---\emph{Added},
\emph{Removed}, or \emph{Modified}---and the judge is asked only whether
this edit set accomplishes the instruction $x$. This is the sense in which
the judge is GT-free; the reference deck is used (if at all) only by a
separate visual-quality judge over rendered images.

\subsection{Identity-based alignment}
Slides are aligned by source id rather than by position or text
similarity: matched ids are paired (absorbing positional shifts), ids new
to $s$ are marked \emph{added}, and ids absent from $s$ are marked
\emph{removed}. Node children are matched by a cascade of strategies---(0)
stable id, (1) unique layer name, (2) text content, (3) type-and-order---so
that generic plugin layer names (\texttt{Caption}, \texttt{Subheader})
still align correctly. Identity-based matching is content-free and
order-preserving, which avoids the false pairings a text-Jaccard matcher
would produce under cross-slide content consolidation. Because pure id
alignment would \emph{hide} reorderings (matched content yields no property
delta), we additionally emit explicit \texttt{slide\_reorder} and per-node
z-order entries when an id-paired element changes position.

\subsection{Compared properties}
Following PPTArena's \texttt{pptx\_to\_json} property set, each aligned node
is compared on: text content (\texttt{characters}); typography
(\texttt{fontFamily}, \texttt{fontSize}, \texttt{fontWeight},
\texttt{fontStyle}, \texttt{textAlignHorizontal}, \texttt{letterSpacing},
\texttt{lineHeight}, \texttt{textCase}, \texttt{textDecoration}); position
and size (\texttt{absoluteBoundingBox}); rotation, opacity, visibility;
fills (RGBA), strokes, effects; and children structure (missing/extra
nodes). We also surface per-character style overrides
(\texttt{characterStyleOverrides} / \texttt{styleOverrideTable}) so that
in-place word-level highlighting is not misread as text deletion, and slide
transitions, which are injected from the full snapshot because the slim
structural export omits them.

\subsection{Normalization and tolerance}
To suppress schema noise and report only semantically meaningful edits, the
diff normalizes equivalent node types ($\textsc{shape\_with\_text}\equiv
\textsc{frame}$) and empty text (\texttt{''}, \texttt{None}, \texttt{`None'}
collapse to empty); rebases each slide subtree to slide-local coordinates;
applies a $\pm 1\%$ relative tolerance to numeric properties (font size,
line height, bounding boxes); and uses a flat absolute tolerance of $0.01$
($\approx\!2/255$, sub-perceptible, matching PPTArena's $\pm 2/255$ hex
tolerance) for $0$--$1$ color channels, where relative tolerance would be
ill-defined near black. Slide-frame $x/y$ are ignored since slide
containers are fixed.

\subsection{Post-processing and scoring}
Raw entries are then (i) collapsed across re-parenting (a node moved to a
new parent FRAME, which the per-parent matcher would otherwise split into a
remove+add pair, is reported as a single \emph{moved} entry); (ii)
consolidated per shape (the four per-axis bounding-box deltas of a
move-and-resize merge into one \texttt{bbox} entry); and (iii)
de-duplicated, with the surviving entry annotated ``applied to $N$
same-named siblings'' so a uniform bulk edit is not mistaken for an isolated
one. A PPTArena-style similarity score $\,1-\tfrac{d}{d+100}\,$ (with $d$
the number of distinct edits) is reported, and the list is capped at 200
entries to bound judge context.

\subsection{Rendering for the judge}
The structured diff is serialized as an indented tree that mirrors the node
hierarchy, so the judge can tell, e.g., that a fill change on a wrapper
FRAME is a table-cell background rather than the text color of a child
node. Each node label is enriched with a short snippet of its descendant
text, which lets the judge identify a semantic entity behind a generic
layer name. Crucially, the diff alone cannot distinguish a slide the agent
\emph{skipped} from one that was \emph{already in the requested state}
(both yield no delta); we therefore append a per-slide snapshot of $D_0$
(text content, alignment, font, and visual shapes) so the judge can verify
whether an untouched slide actually required editing. Finally, rendered
prediction images are attached only when the diff touches visual content
(image/vector/picture-fill additions or modifications), letting the judge
confirm that, e.g., an added picture depicts what the instruction asked
for, while text-only edits are scored from the diff alone to save tokens.

\section{Trajectory Analysis}
\label{app:traj}
To characterize task difficulty without assuming a ground truth, we analyze
the execution trajectories of \texttt{claude-sonnet-4-6} (agentic). The
agent edits through \texttt{batch\_execute}, a single tool call that applies
an ordered array of primitive operations (creation, modification, and
layout) in one commit; collapsing many edits into one batched call avoids
per-operation API round-trips and is the main reason ACE issues few
\emph{agent turns} despite executing many operations. We define
\emph{agent turns} as the number of LLM calls (assistant messages) per task
and \emph{total operations} as the number of primitive commands executed
per task, counting each command inside a \texttt{batch\_execute} call
together with each standalone tool call. Agent turns are right-skewed
(mean 5.6, median 4) with a $7{+}$-turn tail, and total operations have
median 14 / mean 26.1, with $\sim$14\% of tasks exceeding 50, across 56
distinct API types; task difficulty does not track input scale. Per-task node counts and
operation distributions are in Appendix~\ref{app:data}.

\section{Benchmark Details}
\label{app:data}
\begin{figure*}[t]
\centering
\includegraphics[width=\textwidth]{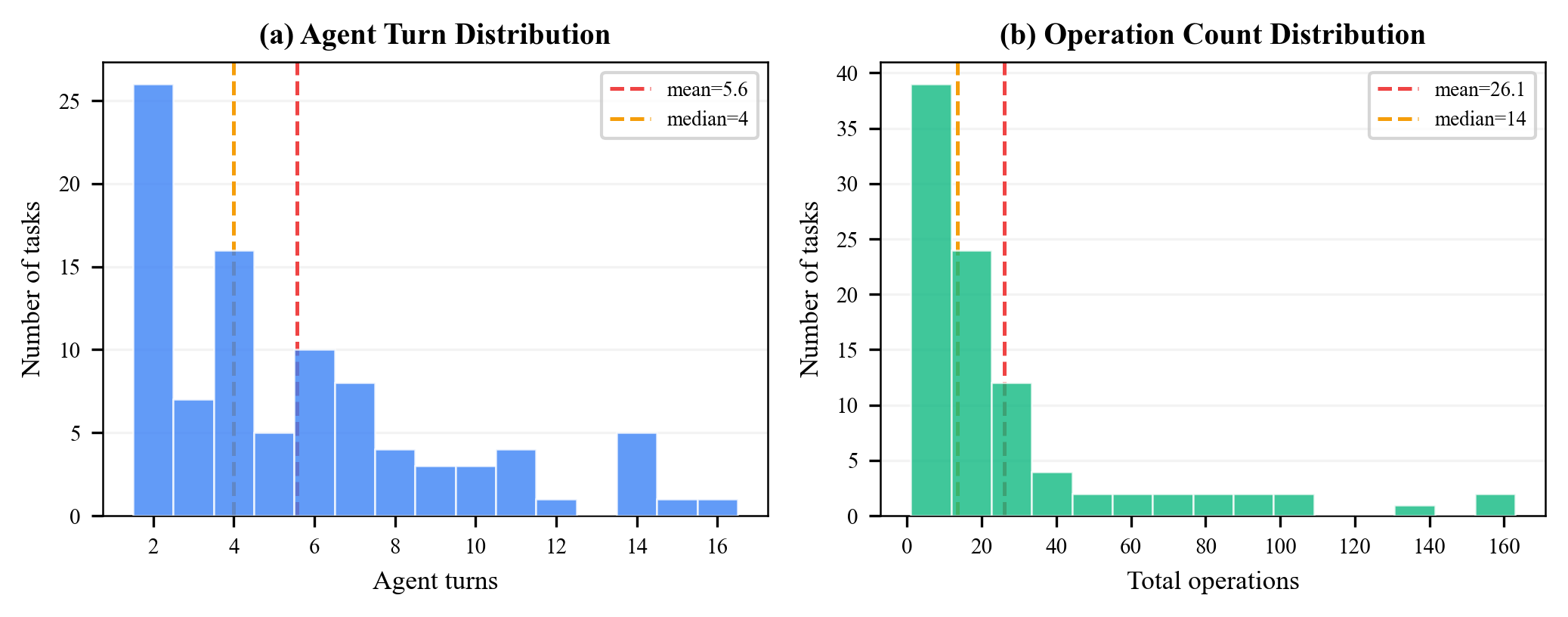}
\caption{Distribution of (a) reasoning turns and (b) total API operations.
Both metrics exhibit long-tail characteristics, ensuring the benchmark
tests both efficiency and endurance in long-horizon editing tasks.}
\label{fig:dist_analysis}
\end{figure*}

\begin{figure}[t]
\centering
\includegraphics[width=\columnwidth]{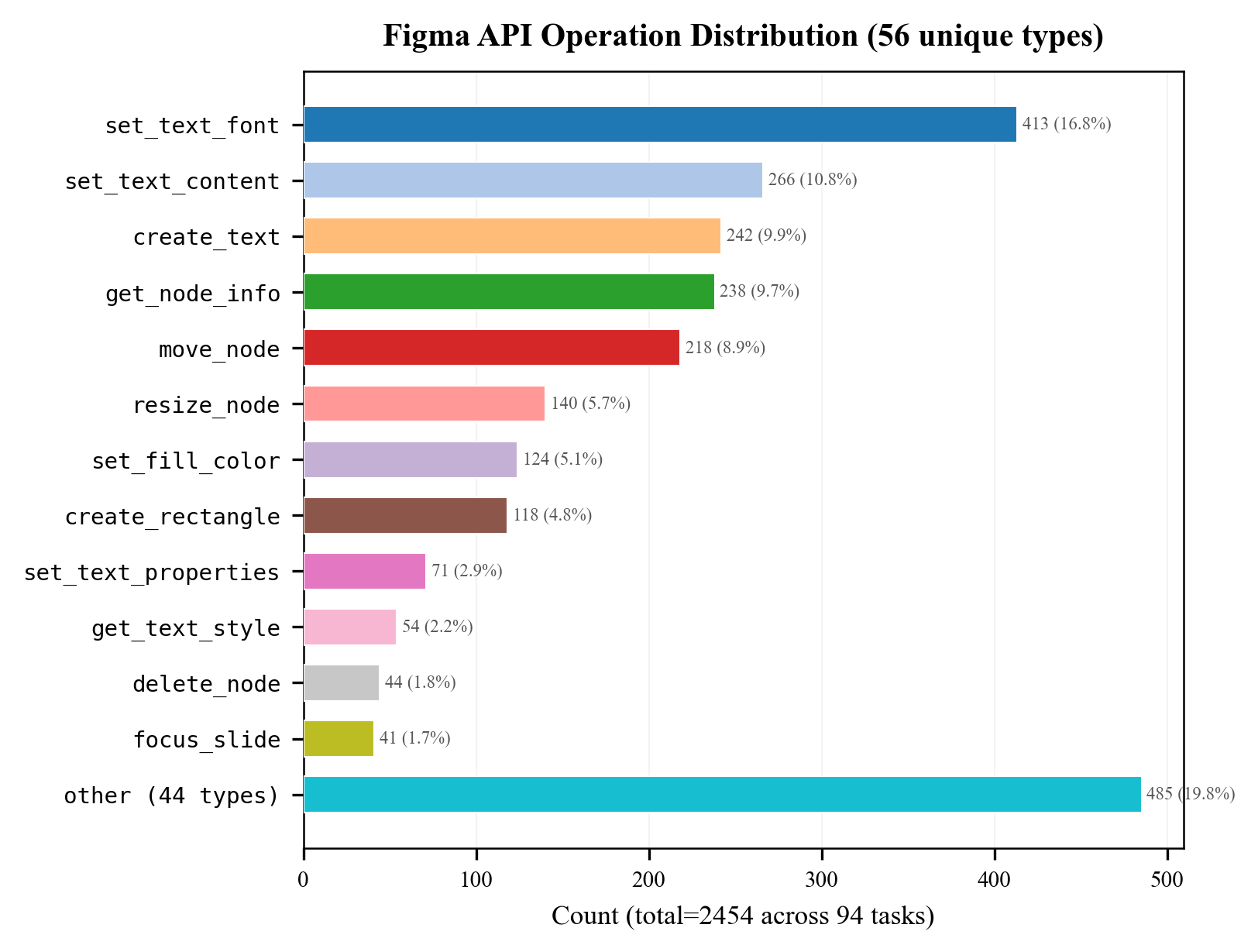}
\caption{Figma API operation distribution. The top 12 operation types
account for the majority of the 2{,}454 total operations, spanning text
styling, spatial layout, and element creation.}
\label{fig:ops_analysis}
\end{figure}

\begin{figure}[t]
\centering
\includegraphics[width=\columnwidth]{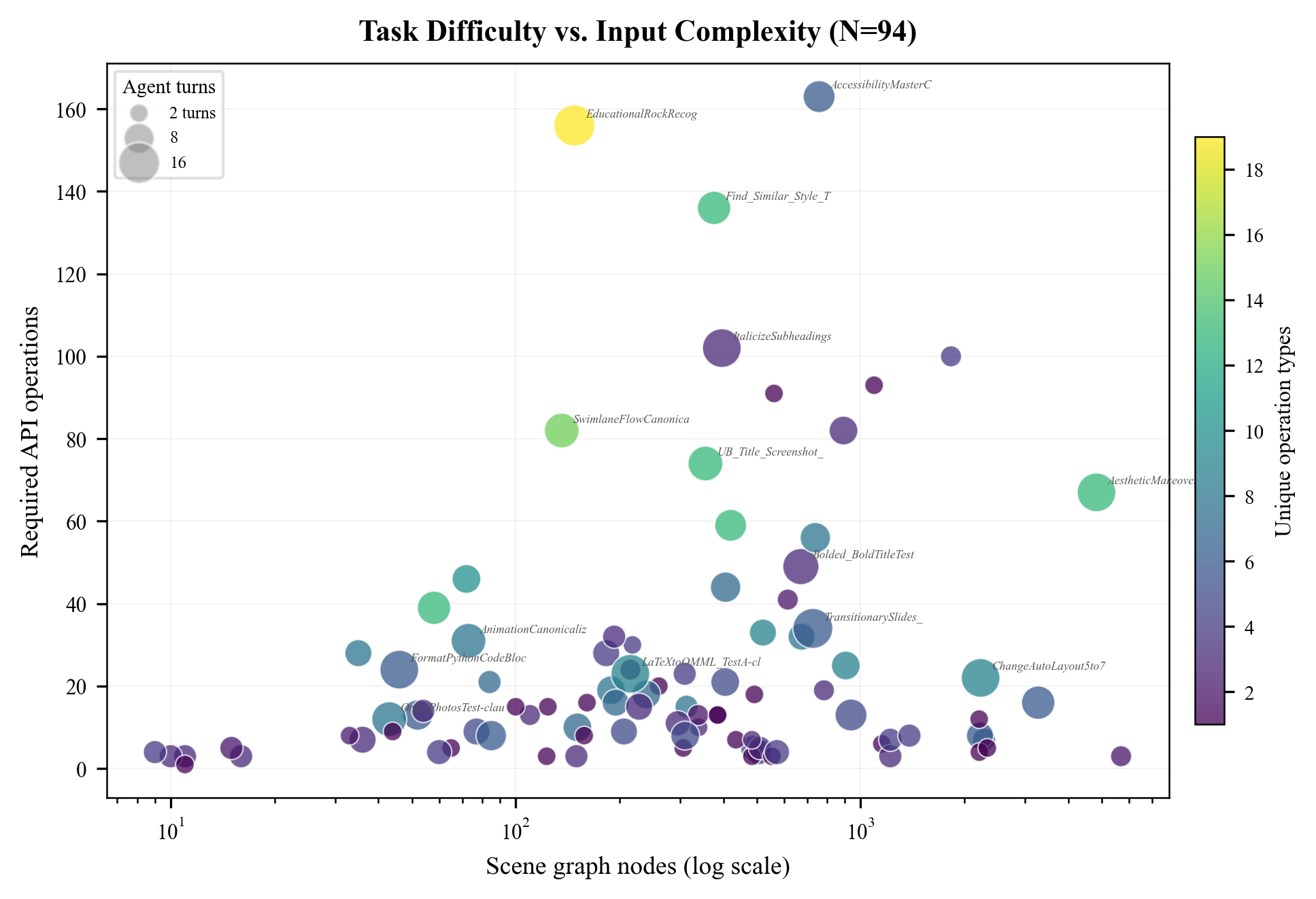}
\caption{Task difficulty vs.\ input complexity. Each point is a task;
x-axis: Figma nodes (log scale), y-axis: total API operations, point size:
agent turns, color: unique operation types. High-density tasks like
\textit{EducationalRockRecognition} (148 nodes, 156 operations) highlight
the benchmark's focus on logic over raw scale.}
\label{fig:complexity_analysis}
\end{figure}

\begin{figure*}[p]
\centering
\includegraphics[height=0.85\textheight, keepaspectratio]{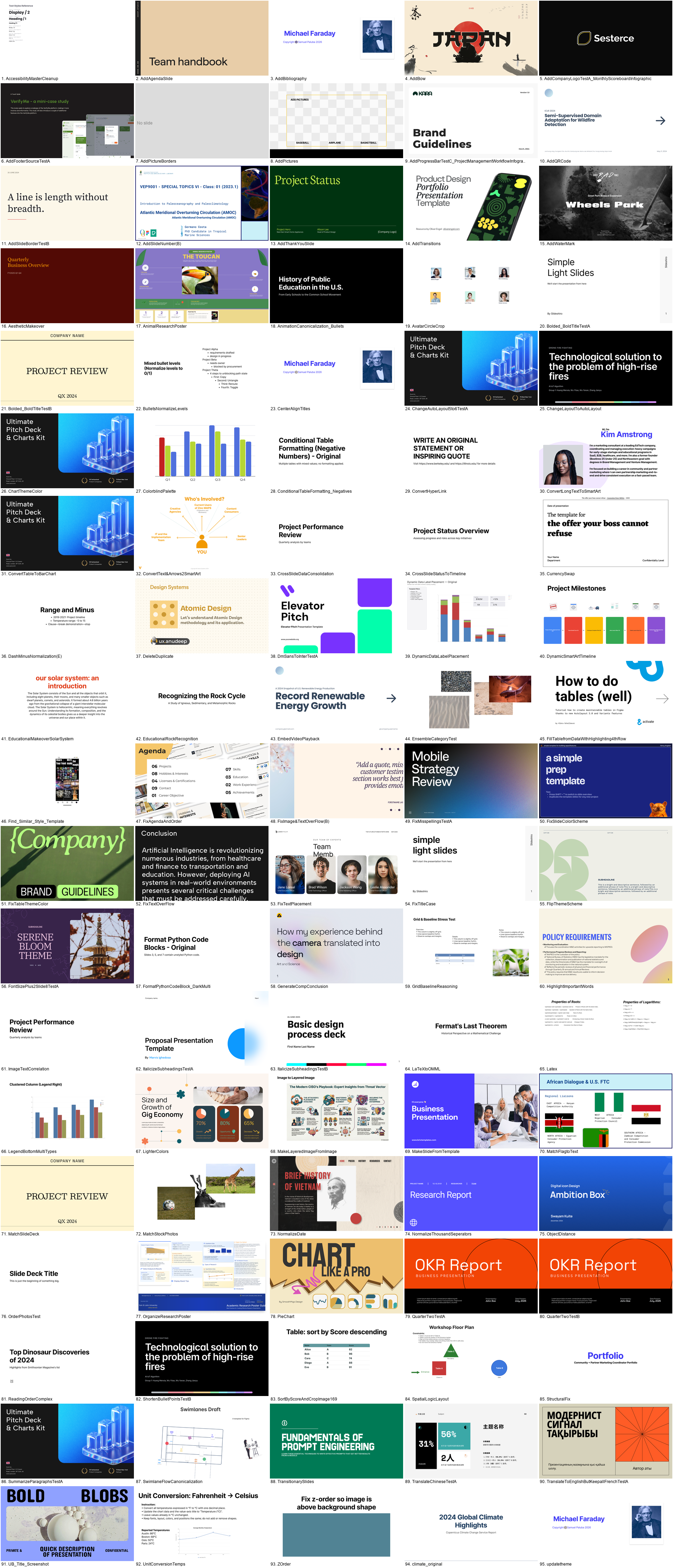}
\caption{Overview of the Figma-Slide benchmark (97 curated tasks). The grid
displays 95 thumbnails: one origin is text-based (no thumbnail), and
Case~112 is omitted because its source thumbnail is identical to that of
Case~105.}
\label{fig:figma_benchmark_overview}
\end{figure*}

All 97 tasks were authored manually. Tables~\ref{tab:removed_cases} and
\ref{tab:novel_tasks} list the 15 removed cases and the 12 novel tasks;
Figures~\ref{fig:dist_analysis}--\ref{fig:figma_benchmark_overview} give
the trajectory, operation, and complexity distributions and the thumbnails
of the full benchmark.

\begin{table*}[t]
\centering\small
\adjustbox{max width=\textwidth}{%
\begin{tabular}{clll}
\toprule
\textbf{Case} & \textbf{Edit Type} & \textbf{Category} & \textbf{Removal Reason} \\
\midrule
13 & Text \& Typography & Content, Styling & Scope too broad \\
18 & Text \& Typography & Content & Duplicate (8, 9) \\
20 & Text \& Typography & Content & Unsupported (notes) \\
26 & Theme \& Background & Styling & Upgraded to novel task \\
27 & Images \& Pictures & Content & Duplicate (16) \\
30 & Charts & Content & Upgraded to novel task \\
34 & Text \& Typography & Content & Upgraded to novel task \\
36 & Slide/Section Mgmt. & Structure & Unsupported (notes) \\
40 & Text \& Typography & Content, Layout & Duplicate (22) \\
41 & Theme \& Background & Styling & Duplicate (33) \\
47 & Charts & Content & Duplicate (29) \\
53 & Tables & Content & Upgraded to novel task \\
85 & Text \& Typography & Content & Duplicate (1, 19) \\
94 & Slide Layout & Layout, Structure & Unsupported (slide size) \\
95 & Template \& Master & Styling, Structure & Unsupported (layout system) \\
\bottomrule
\end{tabular}}
\caption{Summary of the 15 removed cases and their removal reasons.}
\label{tab:removed_cases}
\end{table*}

\begin{table*}[t]
\centering\footnotesize
\renewcommand{\arraystretch}{1.2}
\newcolumntype{C}[1]{>{\centering\arraybackslash}p{#1}}
\newcolumntype{Y}{>{\centering\arraybackslash}X}
\begin{tabularx}{\textwidth}{@{} c C{8cm} Y @{}}
\toprule
\textbf{ID} & \textbf{Simplified Prompt} & \textbf{Design Dimension} \\ \midrule
C1 & Decompose AI-generated image into layered SVG components and create connectors. & Asset Decomposition \\ \addlinespace
C2 & Generate a 9-page presentation based on a template. & Multi-page Synthesis \\ \addlinespace
C3 & Find a suitable template for the reference image. & Style Retrieval \\ \addlinespace
C4 & Create a protein-food infographic from text (green theme). & Creative Synthesis \\ \addlinespace
C5 & Scale AutoLayout (5 to 6) and rearrange contents. & Structural Scaling \\ \addlinespace
C6 & Align list items using the \texttt{AutoLayout} system. & Responsive Layout \\ \addlinespace
C7 & Apply a color palette from other pages to the current slide. & Theme Consistency \\ \addlinespace
C8 & Convert table data into a bar chart with a custom legend. & Data Visualization \\ \addlinespace
C9 & Render raw LaTeX code into visual math notation. & Academic Rendering \\ \addlinespace
C10 & Transform text into SmartArt without overlaps. & Content Vis. \\ \addlinespace
C11 & Fill a table and apply conditional styling (4th row). & Complex Tool-use \\ \addlinespace
C12 & Scale AutoLayout (5 to 7) and rearrange contents. & Structural Scaling \\ \bottomrule
\end{tabularx}
\caption{Specifications for the 12 novel design tasks (Cases 101--112).
Cases C2--C4 (102--104) are released without automatic scores.}
\label{tab:novel_tasks}
\end{table*}

\section{Full Per-Category and Novel-Task Results}
\label{app:fullresults}
Table~\ref{tab:percat} reports ACE's per-category IF/VQ on the 94 evaluable
tasks (\texttt{claude-sonnet-4-6} backbone, \texttt{gpt-5.5} judge,
matching Table~\ref{tab:stats}); Table~\ref{tab:novel_full}
gives the per-task scores for the 9 novel editing tasks.

\begin{table}[h]
\centering\small
\adjustbox{max width=\columnwidth}{%
\begin{tabular}{lrcc}
\toprule
Category & N & IF & VQ \\
\midrule
Content       & 63 & 4.24 & 3.81 \\
Layout        & 35 & 4.26 & 3.54 \\
Styling       & 28 & 4.29 & 3.79 \\
Structure     & 13 & 4.46 & 3.69 \\
Interactivity & \ 4 & 3.50 & 2.75 \\
\midrule
\textbf{All (94)} & \textbf{94} & \textbf{4.23} & \textbf{3.66} \\
\bottomrule
\end{tabular}}
\caption{ACE per-category results on the 94 evaluable tasks. Categories
overlap, so counts sum above 94. The 53-case subset scores 4.45/4.02.}
\label{tab:percat}
\end{table}

\begin{table}[h]
\centering\small
\adjustbox{max width=\columnwidth}{%
\begin{tabular}{r l cc}
\toprule
\# & Novel editing task & IF & VQ \\
\midrule
105 & Auto-layout scale 5$\to$6        & 5 & 5 \\
111 & Table fill + 4th-row highlight   & 5 & 5 \\
106 & Convert to auto-layout           & 4 & 5 \\
112 & Auto-layout scale 5$\to$7        & 5 & 4 \\
108 & Table $\to$ bar chart            & 4 & 4 \\
110 & Long text $\to$ SmartArt         & 5 & 3 \\
107 & Match slide colors to theme      & 4 & 2 \\
101 & Layered-image decomposition      & 2 & 1 \\
109 & LaTeX $\to$ SVG$^{\ddagger}$     & 0 & 0 \\
\midrule
    & \textbf{Mean}                    & \textbf{3.78} & \textbf{3.22} \\
\bottomrule
\end{tabular}}
\caption{ACE (full, with self-correction) on the 9 novel \emph{editing}
tasks with no PowerPoint analogue (1--5 scale; $^{\ddagger}$execution
failure scored 0 via origin-fallback). Seven of nine reach IF$\ge$4; the
open failures are asset decomposition (101) and LaTeX rendering (109).}
\label{tab:novel_full}
\end{table}

\section{Blind Human Study: Protocol and Full Results}
\label{app:human}
\paragraph{Protocol.}
We recruited non-expert raters (lab members and industry colleagues) for a
blind, side-randomized pairwise study on the same outputs the judge
scored. The ACE-vs-HTML head-to-head compares ACE
(\texttt{claude-sonnet-4-6}) against the Claude-Skill HTML baseline with
each
system shown through its own renderer, so raters never penalize
engine-level rendering differences. Raters see only (instruction,
before-render, after-render) and pick win/tie/loss---a relative choice,
which laypeople make more reliably than a 1--5 absolute score; the judge's
scores are hidden, and the judge additionally reads the structural diff (a
different modality), so agreement is a strict test. After excluding two
low-quality raters under pre-stated rules (one all-tie straight-liner; one
with $\ge$70\% one-side position bias), \textbf{26 raters} cast
\textbf{935 judgments}, 13--14 per case, over \textbf{51} ACE-vs-HTML and
\textbf{17} self-correction cases. The counts are 51/17 rather than 53/18
because we drop tasks whose edit is imperceptible in static before/after
renders---a Dissolve slide-transition animation (Case~37) and an
accessibility font adjustment (Case~67, also excluded from the
self-correction set)---since a rater cannot fairly judge an edit they
cannot see. Inter-rater agreement is fair with many ties (Fleiss $\kappa$
0.20--0.29; raw agreement 65--67\%), which we report plainly; the
aggregate preferences below are nonetheless significant.

\paragraph{Judge--human agreement.}
On decided cases---ties excluded on both sides, since a tie carries no
direction to agree on---the in-loop \texttt{gpt-5.5} judge matches the
blind human majority (Table~\ref{tab:human_agree}).

\begin{table}[h]
\centering\small
\adjustbox{max width=\columnwidth}{%
\begin{tabular}{@{}lccc@{}}
\toprule
Dim & Judge--human agreement & $n$ & $p$ vs.\ chance \\
\midrule
IF      & 80\% & 41 & $10^{-4}$ \\
VQ      & 76\% & 37 & .003 \\
Overall & 78\% & 50 & $10^{-4}$ \\
\bottomrule
\end{tabular}}
\caption{Judge--human agreement on decided cases (pooled).}
\label{tab:human_agree}
\end{table}

\paragraph{ACE vs.\ Claude-Skill HTML (same backbone).}
Decisive win-rates (ties dropped): IF 59.6\% [54.5, 64.4] ($p{=}.0003$),
VQ 57.1\% ($p{=}.0025$), Overall 58.7\% ($p{=}.0001$). Every win-rate CI
excludes 0.5 and every preference-mean CI excludes 0.

\paragraph{Self-corrected vs.\ single-pass.}
Table~\ref{tab:human_sc}: blind humans prefer the self-corrected output
$\sim$81\% of the time---including on VQ, which never enters the stopping
signal---so the IF gain is not optimization toward the judge; people
independently see the corrected edits as better.

\begin{table}[h]
\centering\small
\adjustbox{max width=\columnwidth}{%
\begin{tabular}{@{}lccc@{}}
\toprule
Dim & Win-rate [95\% CI] & Binom.\ $p$ (case-level) & Judge$=$human \\
\midrule
IF      & 80.9\% [73.3, 86.7] & $<$0.001 (0.004) & 100\% ($n{=}14$) \\
VQ      & 83.5\% [75.8, 89.0] & $<$0.001 (0.013) & 92\% ($n{=}12$) \\
Overall & 81.5\% [74.1, 87.1] & $<$0.001 (0.013) & 88\% ($n{=}17$) \\
\bottomrule
\end{tabular}}
\caption{Self-correction, blind human preference (17 cases).
Judge$=$human: how often the \texttt{gpt-5.5} judge agrees with the human
majority on decided cases ($n$ is that decided-case count out of 17).}
\label{tab:human_sc}
\end{table}

\section{Out-of-Loop Judges and Same-Backbone Anchor}
\label{app:oojudge}
\paragraph{Self-correction under out-of-loop judges.}
We re-score the identical iteration-1$\to$final pairs of the 18
loop-entering tasks (Table~\ref{tab:oojudge_sc}). The noise floor is the
halted-case control: the 35 tasks whose iteration-1 and final decks are
byte-identical (the complete identical-output subset), so any $\Delta$ is
pure cross-run judge noise. Every judge's gain dwarfs its own floor;
out-of-loop judges recover roughly two-thirds of the in-loop gain,
bounding any critic-specific component at about one-third of the measured
effect.

\begin{table}[h]
\centering\small
\adjustbox{max width=\columnwidth}{%
\begin{tabular}{@{}lcccc@{}}
\toprule
Judge (role) & $\Delta$IF & $\Delta$VQ & impr./unch./degr. & noise floor \\
\midrule
gpt-5.5 (in-loop)        & $+$0.94 & $+$0.78 & 13/2/3 & $+$0.14 \\
claude-4.6 (out-of-loop) & $+$0.61 & $+$0.56 & 9/7/2  & $+$0.03 \\
gemini-3.5 (out-of-loop) & $+$0.56 & $+$1.33 & 9/6/3  & $+$0.17 \\
\bottomrule
\end{tabular}}
\caption{Self-correction gain re-scored out of the loop (18 loop-entering
tasks). Noise floor $=$ mean $\Delta$IF that judge assigns to the 35
byte-identical halted cases.}
\label{tab:oojudge_sc}
\end{table}

\paragraph{Same-backbone anchor vs.\ PPTArena.}
ACE and PPTArena both on \texttt{gpt-5.5}, on the 53-task head-to-head,
instruction following (Table~\ref{tab:anchor}): with the backbone held
fixed and the judge swapped out of the loop, ACE leads the strongest OOXML
baseline by more than 2 IF points, significantly, under every judge.

\begin{table}[h]
\centering\small
\adjustbox{max width=\columnwidth}{%
\begin{tabular}{@{}lccc@{}}
\toprule
Judge & $\Delta$IF [95\% CI] & Wilcoxon & W/T/L \\
\midrule
gpt-5.5    & $+$2.36 [$+$1.77, $+$2.94] & $p{<}0.001$ & 35/15/3 \\
claude-4.6 & $+$2.13 [$+$1.51, $+$2.75] & $p{<}0.001$ & 33/15/5 \\
gemini-3.5 & $+$2.40 [$+$1.75, $+$3.04] & $p{<}0.001$ & 32/20/1 \\
\bottomrule
\end{tabular}}
\caption{ACE $-$ PPTArena, both on \texttt{gpt-5.5} (53 tasks), under all
three judges.}
\label{tab:anchor}
\end{table}

\section{CARE: Quality Ablation and Routing Audit}
\label{app:careaudit}
\begin{table}[h]
\centering\small
\adjustbox{max width=\columnwidth}{%
\begin{tabular}{@{}lccc@{}}
\toprule
 & \multicolumn{3}{c}{$\Delta$IF\,/\,$\Delta$VQ when removed} \\
\cmidrule(l){2-4}
Component (eval set) & claude & gemini & gpt-5.5 \\
\midrule
Repr.\ SG$\to$OOXML (53)              & $-$0.64\,/\,$-$0.62 & $-$0.78\,/\,$-$0.63 & $-$0.69\,/\,$-$0.45 \\
CARE$\to$full context (16)            & $-$0.75\,/\,$-$0.38 & $+$0.00\,/\,$+$0.19 & $-$0.38\,/\,0.00 \\
Spec.\ tools$\to$primitives (16/4/11) & $-$0.56\,/\,$-$0.31 & $-$1.00\,/\,$-$0.75 & $-$0.55\,/\,$-$0.45 \\
Self-correction on$\to$off (53)       & $-$0.41\,/\,$-$0.27 & ---                 & --- \\
\bottomrule
\end{tabular}}
\caption{Leave-one-out component isolation: one component removed, the
other four and the evaluation held fixed (same backbone, \texttt{gpt-5.5}
judge, single pass for the top three rows), scored on the sub-population
where the component is active. Construction and exclusions in
Appendix~\ref{app:careaudit}.}
\label{tab:iso}
\end{table}

\paragraph{Component-isolation construction (Table~\ref{tab:iso}).}
Each row removes exactly one piece with the other four and the evaluation
held fixed, scored only where the component is active. The
CARE$\to$full-context ablation runs on the 16 multi-slide tasks where CARE
actually reduces context, after two principled exclusions: single-slide
tasks (the routed context already \emph{is} the full deck, so the ablation
is a no-op) and 3 large decks (Cases 33/43/57) whose full context exceeds
the 1M-token window without CARE---infeasible to even run in the
full-context condition, itself direct evidence that CARE keeps large decks
runnable. The specialized-tools ablation runs, per backbone, on exactly
the tasks where that backbone's ACE run invoked a
chart/table/SmartArt/math/image tool (claude 16, gemini 4, gpt-5.5 11);
removing a tool a run never called is a no-op that would only dilute the
effect. Removing the tools also inflates operation counts
$\approx$1.8$\times$ (claude 34$\to$59, gpt 28.5$\to$51.2 mean ops; up to
18$\to$127 on construct-heavy tasks), as the same chart or table is
rebuilt from primitives.

\paragraph{Quality ablation.}
On the 16 multi-slide tasks, replacing CARE's routed slice with the entire
deck lowers quality, not just cost (Table~\ref{tab:care_quality}). The
drop is not an overflow artifact (0/16 \texttt{claude} full-context cases
logged an overflow) and is uncorrelated with deck size---a 575K-token deck
drops 0.

\begin{table}[h]
\centering\small
\adjustbox{max width=\columnwidth}{%
\begin{tabular}{@{}lccc@{}}
\toprule
Backbone (16 tasks) & CARE IF/VQ & Full-context IF/VQ & $\Delta$IF \\
\midrule
claude-4.6 & 4.38\,/\,3.88 & 3.62\,/\,3.50 & $-$0.75 \\
gpt-5.5    & 4.62\,/\,3.88 & 4.25\,/\,3.88 & $-$0.38 \\
gemini-3.5 & 3.75\,/\,3.31 & 3.75\,/\,3.50 & $+$0.00 \\
\bottomrule
\end{tabular}}
\caption{CARE quality ablation on the 16 multi-slide tasks where routing
reduces context.}
\label{tab:care_quality}
\end{table}

\paragraph{What each routed representation carries.}
The four representations are complementary, not nested---macro and
systemic each drop what the other keeps (Table~\ref{tab:care_matrix})---so
``correct routing'' means choosing a representation that carries what the
edit needs.

\begin{table}[h]
\centering\scriptsize
\adjustbox{max width=\columnwidth}{%
\begin{tabular}{@{}lcccc@{}}
\toprule
 & micro & macro & systemic & full \\
\midrule
scope & target slide(s) & whole deck & whole deck & whole deck \\
text content & \checkmark & \checkmark & $\times$ & \checkmark \\
typography & \checkmark & \checkmark & \checkmark & \checkmark \\
fill/stroke colour & \checkmark & $\times$ & \checkmark & \checkmark \\
slide background & \checkmark & $\times$ & \checkmark & \checkmark \\
layout / bounding boxes & \checkmark & target only & $\times$ & \checkmark \\
node tree (id/name/type) & \checkmark & \checkmark & $\times$ & \checkmark \\
\bottomrule
\end{tabular}}
\caption{Property coverage of each routed representation. micro $=$ full
scene-graph of the target slide(s); macro $=$ deck-wide skeleton; systemic
$=$ design tokens.}
\label{tab:care_matrix}
\end{table}

\paragraph{Routing audit.}
We measure routing accuracy directly on three axes---two defined from the
instruction alone (no automatic gold required) and one from a structural
diff of the reference (Table~\ref{tab:care_audit}). Verdicts compare the
router's logged mode with the gold mode under the scope ordering micro
$\subset$ \{macro, systemic\} $\subset$ full: \emph{exact} on match;
\emph{over-scope} when the router chose a strictly wider mode (harmless;
costs only tokens); \emph{under-scope} when it chose a strictly narrower
mode (the one failure that can drop needed context); \emph{borderline}
when several representations are each sufficient and the router picks one
of them. Over the 53 tasks: 47 exact $+$ 5 borderline $+$ 1 under-scope
$+$ 0 over-scope. The 5 borderline are add-a-slide or mixed
theme-plus-edit tasks where macro/systemic/full are each defensible (e.g.\
``Add a Thank-you slide''); a disputed borderline label can only move
between exact and borderline, never into under-scope, so the 52/53 is
robust to relabeling.

\begin{table}[h]
\centering\scriptsize
\adjustbox{max width=\columnwidth}{%
\begin{tabular}{@{}p{1.9cm}p{2.0cm}p{3.1cm}@{}}
\toprule
Axis & Result & Measured how \\
\midrule
Mode accuracy & 52/53 (47 exact $+$ 5 borderline; 1 under-, 0 over-scope)
& gold mode $=$ minimal sufficient representation, labelled from
instruction semantics against Table~\ref{tab:care_matrix} (no diff) \\
Slide-selection recall (micro) & 1.00 (16/16; 20/20 all modes) & router
targets vs.\ \textsc{diff}(original, reference), on the 20 tasks whose
original and reference share node IDs \\
Explicit-slide accuracy & 13/13 & instructions naming a slide number are
their own gold \\
\bottomrule
\end{tabular}}
\caption{CARE routing audit on the 53-task subset.}
\label{tab:care_audit}
\end{table}

\paragraph{Misrouting is bounded and recoverable.}
The errors are one-sided: over-scoping never occurs and would cost only
tokens, so the sole quality-relevant failure is under-scoping, of which
there is exactly 1/53---Case~75, a deck-wide grid/alignment cleanup routed
to macro when raw per-slide geometry was needed (IF 3.0 vs.\ the 4.04 deck
mean), which self-correction lifts back to 4.0. A miss is also cheap,
because CARE reduces only the \emph{initial} context and never blinds the
agent: live inspection tools (\texttt{get\_node\_info},
\texttt{get\_all\_slides}) return full node data at execution time.
Case~24 demonstrates the recovery---routed to macro (which carries no
colour and, here, no image), the agent still produced an agenda slide
matching the deck background by issuing 14 \texttt{get\_node\_info} calls
to read the existing fills before creating the slide. A reduced or
mis-scoped context defers information to a live query; it does not lose
information the edit needs.

\section{Rollback and Loop-Knob Sensitivity}
\label{app:sens}
\paragraph{Strict-peak rollback.}
Of the 18/53 tasks entering correction, 3 (17\%) regress in IF, each by
$\le$1 point. Because the critic's score is logged at every iteration, we
keep the last iteration unless the critic's own score declined from an
earlier one, in which case we return that earlier iteration. The rule uses
only the in-loop IF critic---no ground truth, hence deployable as-is---and
yields a strict, side-effect-free gain: IF 4.45$\to$4.49, VQ
4.02$\to$4.06, with no task harmed. (An oracle VQ-aware selector reaches
4.51/4.08 but requires ground truth; we report it only as an upper bound.)

\paragraph{Sensitivity to $\tau$ and $K$.}
Both loop knobs are replayed from logged trajectories (external IF, 53
tasks; the small offset from the reported 4.45 is within the $\le{+}0.17$
re-judging noise bound of Appendix~\ref{app:oojudge}). Both curves in
Table~\ref{tab:sens} are monotone with diminishing returns and a plateau:
in $K$ the marginal gain shrinks ($+0.22$ then $+0.12$), so roughly 65\%
of the benefit arrives by $K{=}2$ though a third iteration still helps; in
$\tau$ the gain plateaus by 3.5, so $\tau{=}4$ sits on a plateau rather
than a lucky sweet spot---lowering $\tau$ degrades gracefully toward the
no-correction baseline (4.04) with no cliff, and raising $\tau$ beyond 4
is not simulable within the generated budget. VQ follows the same shape
(3.75 $\to$ 3.94 $\to$ 4.02 in $K$).

\begin{table}[h]
\centering\small
\adjustbox{max width=\columnwidth}{%
\begin{tabular}{@{}lc@{\hskip 16pt}lc@{}}
\toprule
Max-iter $K$ ($\tau{=}4$) & IF & Threshold $\tau$ ($K{=}3$) & IF \\
\midrule
$K{=}1$ & 4.04 & $\tau{=}2.0$ & 4.04 \\
$K{=}2$ & 4.26 & $\tau{=}3.0$ & 4.19 \\
$K{=}3$ & 4.38 & $\tau{=}3.5$ & 4.38 \\
        &      & $\tau{=}4.0$ (default) & 4.38 \\
\bottomrule
\end{tabular}}
\caption{Loop-knob sensitivity, replayed from logged trajectories
(external IF, 53 tasks).}
\label{tab:sens}
\end{table}

\section{PPTArena: Attempt Rate vs.\ Conditional Quality}
\label{app:pptarena}
The blended 53-task average in Table~\ref{tab:main} conflates ``how often
it attempts'' with ``quality when it does,'' so we decompose it. A case
counts as \emph{edited} iff PPTArena produced a modified deck: it edits
\textbf{33/53 (62.3\%)} and produces no edit on \textbf{20/53
(37.7\%)}---do-nothing punts plus one generation failure. On IF the
do-nothing fallback is effectively zero (mean 0.00--0.15 across the three
judges; Table~\ref{tab:pptarena_decomp}), so the headline does not flatter
the baseline on that axis; only VQ uses the origin-vs-reference
comparison, which is needed to distinguish ``did nothing'' from
``destroyed the deck,'' and that floor is what drags the blended average
down. Even at its conditional quality on the 33 edited tasks, PPTArena
trails ACE (ACE$_{\text{gpt}}$ 4.74/4.19; ACE$_{\text{claude}}$
4.45/4.02), and the same-backbone anchor is $+2.36$ IF at $p{<}0.001$
(Appendix~\ref{app:oojudge}).

\begin{table}[h]
\centering\small
\adjustbox{max width=\columnwidth}{%
\begin{tabular}{@{}lccc@{}}
\toprule
Judge & Attempt rate & Cond.\ (33 edited) IF/VQ & Unedited-20 IF/VQ \\
\midrule
gpt-5.5    & 33/53 (62.3\%) & 3.73\,/\,2.79 & 0.15\,/\,1.40 \\
claude-4.6 & 62.3\%         & 3.85\,/\,2.82 & 0.05\,/\,1.15 \\
gemini-3.5 & 62.3\%         & 3.85\,/\,3.03 & 0.00\,/\,0.95 \\
\bottomrule
\end{tabular}}
\caption{PPTArena decomposition: attempt rate and conditional quality
(blended 53-task averages appear in Table~\ref{tab:main} and
Table~\ref{tab:oojudge}).}
\label{tab:pptarena_decomp}
\end{table}


\section{Qualitative Comparisons}
\label{app:qual}
Figure~\ref{fig:qual_appendix} (page~1) shows five further tasks (beyond
Case~17 in the main text) spanning table edits, layout, multi-element
composition, and structured edits. Each row gives the manually authored
reference and the three pipelines; ACE's edits frequently differ from the
reference while remaining valid.

\section{Tool-Execution Efficiency}
\label{app:tooleff}
\begin{figure*}[t]
\centering
\begin{subfigure}{0.48\textwidth}
  \centering
  \includegraphics[width=\textwidth]{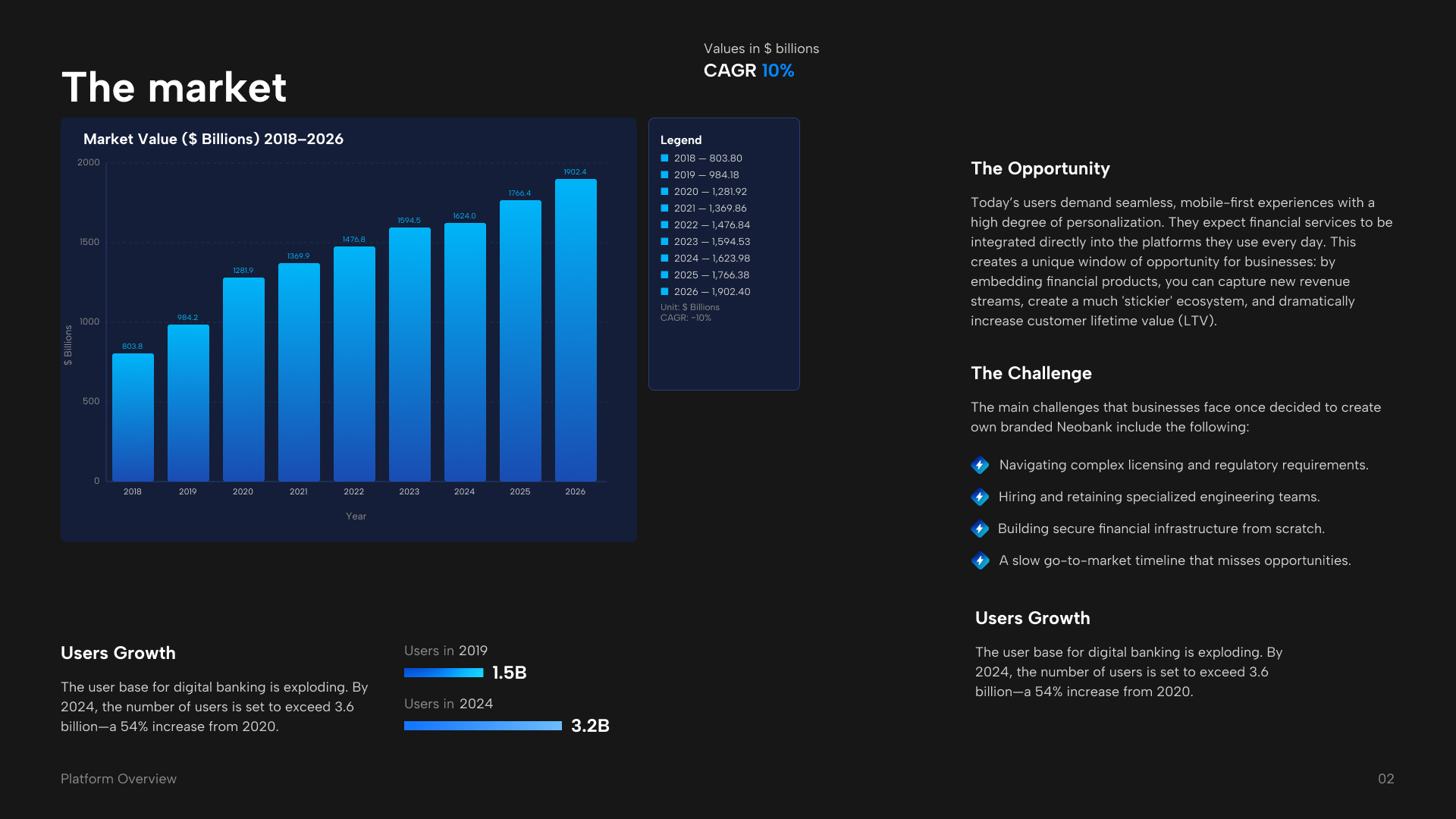}
  \caption{Ours: \texttt{create\_graphics} (22 ops).}
\end{subfigure}
\hfill
\begin{subfigure}{0.48\textwidth}
  \centering
  \includegraphics[width=\textwidth]{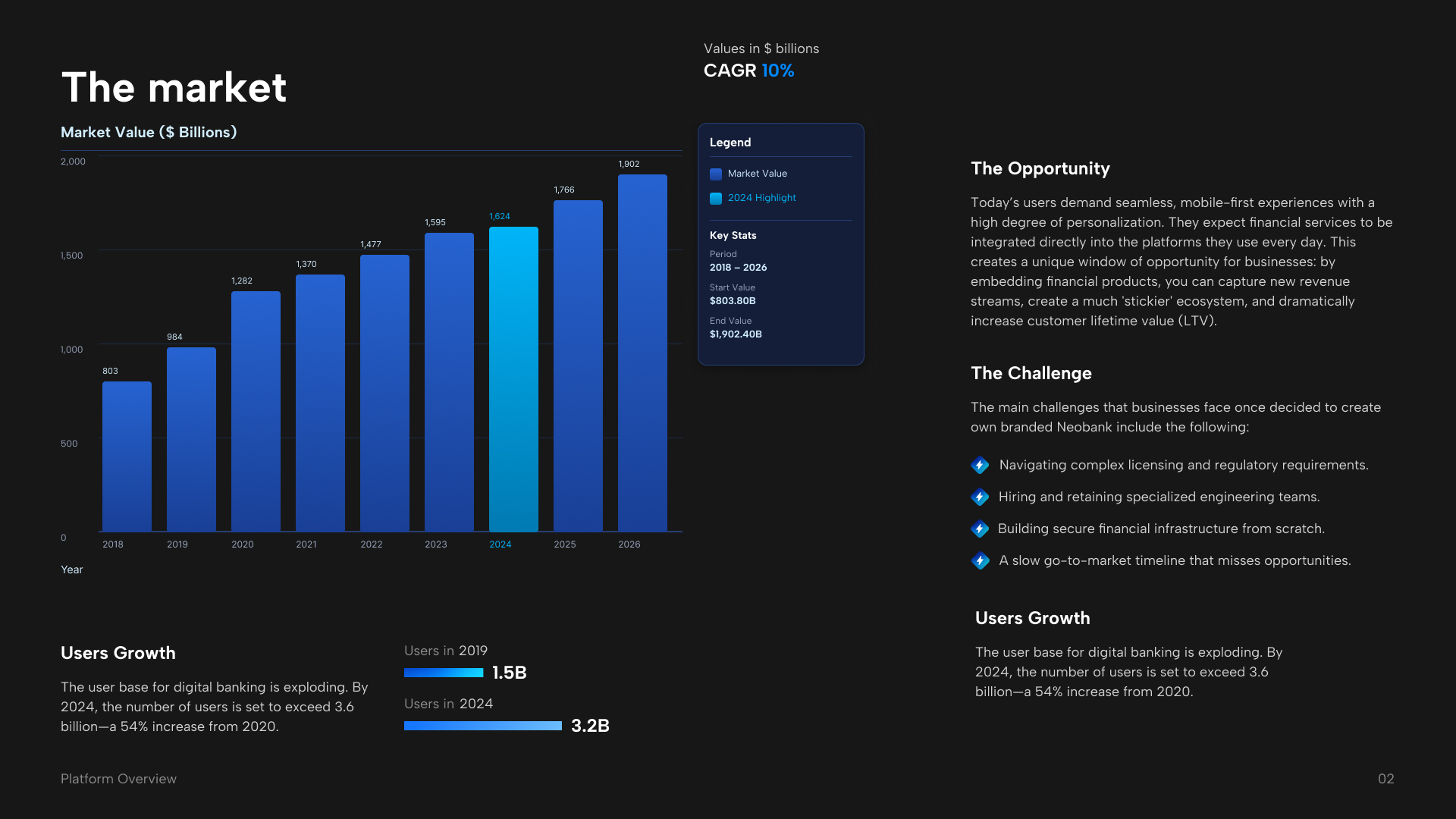}
  \caption{Ablation: primitive tools (66 ops).}
\end{subfigure}
\caption{Tool-execution efficiency on the same ``Table-to-Chart'' task.
(a) The specialized \texttt{create\_graphics} tool needs only 22
operations. (b) Forcing atomic primitives (\texttt{create\_rectangle},
etc.) yields a \textbf{3.0$\times$} increase (66 operations), inflating the
``reasoning tax'' and the chance of alignment errors.}
\label{fig:tool_efficiency_compare}
\end{figure*}

Figure~\ref{fig:tool_efficiency_compare} contrasts the specialized
\texttt{create\_graphics} tool against atomic primitives on the same
``Table-to-Chart'' task: the specialized tool needs only 22 operations
where the primitive route needs 66 (a 3.0$\times$ increase).

\section{Claude-Skill HTML Baseline Construction}
\label{app:html}
The Claude-Skill HTML agent is Claude Code invoked headlessly (\texttt{claude -p})
with file-system tools (read/edit/grep/bash) and a single packaged skill,
\texttt{slide-editor}, that scopes its behaviour to the
deck-as-one-\texttt{index.html} representation. The agent receives the
invocation prompt in Listing~\ref{lst:html-invoke}, which points it at
\texttt{task.txt}, \texttt{index.html}, and the reference renders in
\texttt{frames/}, and requires it to emit the full edited document as
\texttt{output.html}. The skill definition itself
(Listing~\ref{lst:slide-editor}) supplies the operative knowledge: the
slide schema (absolutely-positioned children inside fixed
$1920{\times}1080$ \texttt{.slide} divs, z-order $=$ DOM order), the stable
\texttt{data-slide-id}/\texttt{data-id} handles that mirror Figma node ids
and are the recommended Grep targets (a deck is typically 50K--100K tokens
and cannot be read whole), the output contract (untouched slides stay
byte-identical, CSS inline), and editing principles that protect round-trip
fidelity (minimal edits, no unrequested redesign, and the semicolon-prefix
rule that prevents a newly appended inline property from silently voiding
the preceding declaration). This skill is the HTML-baseline analogue of the
CARE router and executor prompts used by our system. To verify render
fidelity we compute per-page DINOv2 cosine similarity between the HTML
capture and the Figma reference and iteratively repair pages below 0.8;
the resulting captures match the references closely (91.9\% of pages
$\ge$0.95).

\begin{lstlisting}[style=prompt,caption={Invocation prompt passed to \texttt{claude -p} for the Claude-Skill HTML baseline.},label={lst:html-invoke}]
Edit the slide deck in this directory.
- The task is described in task.txt.
- The deck is in index.html (each <div class='slide'> is one slide).
- Reference PNGs for each slide are in frames/slide_<index>.png.
- Apply only the edits requested by the task; leave everything else unchanged.
- Save the complete edited HTML as output.html in this same directory.
Use the slide-editor skill.
\end{lstlisting}

\begin{lstlisting}[style=prompt,caption={\texttt{SKILL.md} for the \texttt{slide-editor} skill used by the Claude-Skill HTML baseline (abridged).},label={lst:slide-editor},float=*]
---
name: slide-editor
description: Edit a slide deck represented as a single HTML file. Use whenever the working directory contains an index.html that visualizes a slide deck (each slide is a div with class="slide", absolute positioning, fixed 1920x1080 size) and the user asks to modify it. Always edit index.html and save the result as output.html in the same directory.
---

# slide-editor

## Input
- index.html  - the deck. Each slide is one <div class="slide" style="width:1920px;height:1080px;...">. Z-order = DOM order.
- images/     - referenced images.
- frames/slide_<index>.png (optional) - pre-rendered visual reference per slide.
- task.txt    - the natural-language editing task.

## Slide schema
- Each .slide: position:relative; width:1920px; height:1080px; overflow:hidden. Children absolutely positioned.
- Text: <div> with left/top/width/height and nested spans for per-character styles.
- Shape: <div> with background-color / border / border-radius. Vector: inline <svg>. Image: background-image url.
- Editable attrs: color, background-color, font-*, letter-spacing, line-height, text-align, left/top/width/height,
  transform:rotate, opacity, border-radius, box-shadow, filter.

## Stable identifiers
Every slide/element carries a stable id (matches Figma node ids):
  <div class="slide" data-slide-id="1:29">, <div data-id="1:30">, <svg data-id="1:31"> ...
ALWAYS prefer Grep on data-slide-id / data-id to navigate. index.html is 50K-100K tokens and CANNOT be Read whole.

## Output contract
- Write the edited HTML to output.html; do NOT modify index.html.
- Copy <!doctype html>, <head>, <style> as-is unless the task requires changes.
- Keep all .slide divs in order; untouched slides must be byte-identical. Keep image paths.

## Working principles
- Edit only the relevant slides; preserve everything not asked (round-trip fidelity is evaluated).
- No design improvements / recoloring / restyling unless requested. Keep CSS inline on elements.
- ALWAYS separate CSS declarations with ';'. Inline values may lack a trailing ';' (e.g. color: rgba(0,0,0,1.0)").
  When appending a NEW property, write "; <prop>: <val>". Otherwise it merges into the previous declaration and the
  WHOLE declaration is dropped as invalid (most common failure: font-style: italic silently lost).

## Workflow
1. Read task.txt.
2. Locate targets WITHOUT reading index.html whole: Grep class="slide"/data-slide-id for boundaries; Grep a task-keyed
   pattern; Read only relevant line ranges (one slide ~25-40 lines). For bulk edits, find ONE pattern and edit via regex.
3. (Optional) Inspect frames/ PNGs for visual grounding.
4. Make the minimal edits, targeting elements by data-id.
5. Save the complete edited HTML to output.html.
6. Verify by Reading back only the changed lines or Grepping changed data-ids. Do NOT re-read the whole deck.
\end{lstlisting}

\section{Prompts}
\label{app:prompts}
This appendix lists the verbatim system prompts used in our pipeline.
Runtime placeholders are written as \texttt{\{name\}} (e.g.\
\texttt{\{instruction\}}, \texttt{\{slideCount\}},
\texttt{\{baseJsonString\}}) and are substituted at inference time. The ACE
design agent is shown in its default \emph{batch} tool-calling
configuration.

\subsection{ACE Design Agent}
\label{app:ace-agent}
The agent prompt is assembled from a fixed \emph{Context} header, a shared
set of \emph{Tool Use Principles}, the \emph{Figma Slides Basics}
reference, an optional context-mode block selected by the CARE router
(\S\ref{app:ace-modes}), the serialized document state, and the user
\emph{Instruction}. Listing~\ref{lst:ace-template} shows the top-level
template (the FULL-context variant); the reusable blocks follow.

\begin{lstlisting}[style=promptstyle,caption={ACE design agent prompt template (FULL context variant).},label={lst:ace-template}]
**Context**
You are a presentation-design agent with access to Figma Slides via tool calls.
Follow the **Instruction** to modify the presentation.
Refer to the **Tool Use Principles** and **Figma Slides Basics** for guidance.

**Tool Use Principles**
{tool_use_principles}

**Figma Slides Basics**
{figma_slides_basics}

**Current Presentation State**
The presentation has {slideCount} slide(s).
Below is the full document structure (JSON) describing every slide and its children.
You already have all node IDs, types, and properties - proceed directly to modifications without calling discovery tools.
```json
{baseJsonString}
```

**Instruction**
{instruction}
\end{lstlisting}

\begin{lstlisting}[style=promptstyle,caption={\texttt{\{tool\_use\_principles\}} block (batch mode).},label={lst:ace-principles},float=*]
Your task is to produce an array of tool (function) calls necessary to modify the presentation in one turn.
Include every necessary tool (function) calls and do not output any text other than the function calls themselves.

1. Plan first - briefly outline the key steps you will take.
2. Be exhaustive - consider all parameters, options, ordering, and dependencies necessary to make the modification in one turn.
3. Preserve existing content - do NOT delete or recreate elements that should remain unchanged. Only modify what the instruction asks for.
4. Use batch_execute aggressively - it supports ALL commands (creation, modification, layout).
   ALWAYS prefer ONE batch_execute call over multiple individual tool calls. Each individual tool call costs a full API round-trip.
   CRITICAL: If you need to create or modify 3+ elements, ALWAYS use batch_execute.
   Example - creating multiple elements + setting properties in one call:
     batch_execute({ operations: [
       { command: "create_frame", params: { x: 0, y: 0, width: 400, height: 40, name: "Row 1", parentId: "1:10" } },
       { command: "create_text", params: { x: 10, y: 5, width: 380, text: "Hello", fontSize: 16, parentId: "1:10" } },
       { command: "set_text_font", params: { nodeId: "1:23", fontFamily: "Inter", fontStyle: "Bold" } },
       { command: "set_fill_color", params: { nodeId: "1:67", color: "#FF0000" } }
     ] })
5. Text layout quality
   - ALWAYS set an explicit **width** on text nodes to prevent overflow/overlap.
   - Predict the bounding box of every element BEFORE placing it. For text nodes, rendered height ~= ceil(text_length_px / width) x fontSize x 1.3 (line-height). Ensure NO overlap with any neighbor in all directions.
   - For multi-column layouts, column width = (available width) / num_columns.
6. Avoid redundant discovery
   - NEVER call check_connection_status - the connection is already established.
   - NEVER call export_json or get_page_structure - the structure is already provided.
   - Do NOT call get_all_slides if slide IDs are already provided.
   - Only use get_node_info / get_node_info_by_types for details not visible in the provided JSON.
   - After modifications, do NOT make extra verification calls - tool results already confirm success/failure.
7. Minimize API round-trips
   - Collect ALL changes across ALL slides into a SINGLE batch_execute when possible.
   - Do NOT alternate focus_slide -> batch_execute per slide.
   - focus_slide is only needed before creating new nodes; existing-node modifications work by nodeId regardless of focus.
8. Handle node ID staleness
   - After deleting a slide/node, ALL child node IDs within it become invalid immediately.
   - NEVER batch a delete_slide with operations referencing nodes on other slides - the delete may invalidate IDs mid-batch.
   - NEVER batch a create_slide with its child element operations - the new slide's ID is unknown until create_slide returns.
   - CRITICAL - create_slide is NEVER the final step. After creating slides, ALWAYS follow up with batch_execute to populate them. A blank slide is never acceptable.
   - CRITICAL - parentId after create_slide: when adding elements to a newly created slide, include parentId (the new slide's ID) in EVERY create_* params; otherwise elements land on the previously focused slide.
   - NEVER batch a create_slide with reorder_slides - instead create all slides first, then reorder_slides with the complete list.
9. Deliver - respond with the exact sequence of tool (function) calls to run.
\end{lstlisting}

\begin{lstlisting}[style=promptstyle,caption={\texttt{\{figma\_slides\_basics\}} block.},label={lst:ace-basics}]
1. Figma Slides Basics
- Each slide is a top-level container (like a frame) holding content elements, arranged in a grid; each slide has a unique ID.
- When creating/modifying content, parentId should be a slide ID.

2. Node Hierarchy
- All content nodes live within slides. Parent-child links form the hierarchy.
- Coordinates in relativeTransform and move_node are relative to the PARENT node, not the slide root (0,0 = top-left of the parent frame).

3. Content Elements
- text, rectangles, ellipses, frames, groups, graphics (SVG), charts, tables.
- Text nodes: font family/style/size/color/alignment. Shapes: fill, stroke, corner radius, opacity.

4. Slide Management
- focus_slide navigates the viewport; ONLY required before creating new nodes on that slide.
- For modifying existing nodes, use batch_execute with nodeId directly - no focus_slide needed.

5. Slide Ordering
- Presentation order is the internal children array, NOT x/y coordinates. move_node does NOT change slide order.
- To reorder, use reorder_slides with the full list of slide IDs in order.

6. Deck theme consistency (new slides/blocks only)
- Match the deck's existing styling; do NOT default to white background + black text unless a comparable slide does.
- Before populating a new slide, inspect 1-2 comparable slides for background fill, title font/size/style/color, and recurring decorations, and copy those exact values.
- For chart/theme recolor: use replace_color with exact source/target hex; recolor only chart/data-series elements.
- Do NOT restyle existing slides/nodes unless explicitly asked.
\end{lstlisting}

\subsubsection{CARE Context-Mode Blocks}
\label{app:ace-modes}
The CARE router selects one context-mode block, inserted after the
\emph{Figma Slides Basics} block, which determines how the document state
is serialized (full scene graph, text-only skeleton, or style-token
summary). Listings~\ref{lst:ace-micro}--\ref{lst:ace-off} give the four
variants.

\begin{lstlisting}[style=promptstyle,caption={MICRO\_SPATIAL --- deep scene graph for target slides only.},label={lst:ace-micro}]
**Context Mode: Micro-Spatial (Targeted Editing)**
- Below is the FULL scene graph for the target slide(s) only.
- Other slides are summarised minimally (ID and name only).
- The full document has already been imported into Figma - all nodes exist.
- Use the node IDs from the scene graph to reference existing elements directly via batch_execute.

**Coordinate System**
- Each node's relativeTransform gives its position relative to its PARENT frame: [[a,b,tx],[c,d,ty]].
- move_node sets PARENT-relative coordinates (same as relativeTransform tx/ty). Derive move_node x/y from relativeTransform, NOT absoluteBoundingBox.
- size gives width/height; use resize_node to change dimensions.
\end{lstlisting}

\begin{lstlisting}[style=promptstyle,caption={MACRO\_PROGRAMMATIC --- text + ID skeleton for batch operations.},label={lst:ace-macro}]
**Context Mode: Macro-Programmatic (Batch Operations)**
- Below is a SKELETON view showing only text content and node IDs (colors/fills/strokes stripped for efficiency).
- Target slides include a bounds field per node: [x, y, width, height] in absolute coordinates (reason about layout, overlap, etc.).
- Identify ALL matching nodes from the skeleton and generate tool calls for EVERY node that needs modification.
- The full document has been imported - reference any node by ID via batch_execute.
\end{lstlisting}

\begin{lstlisting}[style=promptstyle,caption={SYSTEMIC\_TOKEN --- design-token (color/font) summary.},label={lst:ace-systemic}]
**Context Mode: Systemic (Design Token Operations)**
- Below is a style metadata summary of all colours, fonts, and their usage across slides.
- Use replace_color for bulk colour changes, set_fill_color for targeted fills, set_text_font for font changes.
- Target by colour value or font family rather than enumerating individual nodes when possible.
- The full document has been imported - all nodes referenced by ID.
\end{lstlisting}

\begin{lstlisting}[style=promptstyle,caption={OFF --- no pre-supplied context; agent discovers state via tools.},label={lst:ace-off}]
**Presentation Overview**
The presentation has {slideCount} slide(s) and has already been imported into Figma. No document structure is provided.
- Use get_all_slides to find slide IDs, then focus_slide to select one.
- Use get_node_info_by_types or get_text_node_info to inspect node details.
- Use export_slide_image only if visual context is essential.
- Do NOT call check_connection_status, export_json, or get_page_structure.
\end{lstlisting}

\subsection{Instruction-Following (IF) Judge}
\label{app:if-judge}
The IF judge scores whether the agent's edits accomplish the instruction.
It receives a focused diff between the \emph{initial} state and the agent's
\emph{prediction} (plus rendered prediction images when the diff touches
visual content), and emits a single 0--5 score with a one-sentence
justification. Listing~\ref{lst:if-system} gives the system prompt and
Listing~\ref{lst:if-user} the user prompt.

\begin{lstlisting}[style=promptstyle,caption={IF judge --- system prompt.},label={lst:if-system},float=*]
You are a strict judge of INSTRUCTION FOLLOWING for Figma slide editing tasks.

CRITICAL UNDERSTANDING:
- The "Instruction" is what the model/editor received (the user's request).
- You receive a FOCUSED DIFF showing the AGENT'S ACTUAL EDITS - the difference between the INITIAL slide state and the PREDICTION. (No diff against ground truth; GT is checked visually by a separate visual-quality judge.)
- Judge whether those edits accomplish the Instruction.
- "Removed" = elements deleted by the agent. "Added" = elements created. "Modified" = kept but changed.
- An EMPTY/near-empty diff almost always means the agent did NOT perform the instruction (low score).

INITIAL STATE SNAPSHOT (per-slide content, included BELOW the diff):
- For "apply X to every slide", a slide with no diff entry can mean (a) the agent skipped it, or (b) it was ALREADY in the requested state. The snapshot lists each slide's TEXT + alignment + font and visual shapes - use it to distinguish.
- Do NOT treat a missing diff entry as failure unless the snapshot confirms the slide actually needed editing.

FLEXIBILITY:
- Accept different valid approaches. Exact positions/sizes don't matter unless the Instruction requires them.
- Small measurement variations (+/-1%) are acceptable. Focus on semantic properties: text, fonts, colors, structure.

STRUCTURED VISUALS (IMPORTANT):
- Figma Slides has NO native table/list/chart/SmartArt type; these are built from FRAME containers in a grid, with nested FRAMEs/RECTANGLEs as cells and TEXT nodes as content.
- When the Instruction targets a "table"/"chart"/"diagram"/"list", edits to the constituent FRAMEs/RECTANGLEs/TEXTs ARE the relevant edits. A fill-color change on a FRAME nested in the table wrapper IS a cell color change.
- Layer names are arbitrary; judge a node's role by its path position and whether the same edit applies to its siblings.

REORDER / SORT OPERATIONS (IMPORTANT):
- "sort table rows" / "reorder slides" is implemented by swapping sibling indices (zOrder) and y/x positions on many shapes at once. MANY zOrder + boundingBox.y entries on the same parent are ONE semantic operation.
- A children.reorder summary entry IS the sort operation; read it with per-row zOrder entries to verify the resulting order matches the instruction.
- A FRAME tagged "[IMAGE fill]" IS a picture; moving it z-wise IS moving the picture. For "bring X to front / send Y to back", judge by the FINAL stacking, not which shapes moved.
- Do NOT score this as "only z-order changed" - it is a meaningful structural edit.

PRED SLIDE IMAGES (when included):
- You also receive RENDERED IMAGES of the prediction slides.
- Use them to judge content-relatedness the diff cannot show (does an added IMAGE depict what was asked? does positioning match intent?).
- The pred images are NOT a ground-truth target - never compare to a GT image or score on stylistic similarity.

HARSH SCORING POLICY (very strict):
- Choose the lower score when uncertain. For translation/summarization, semantic similarity matters more than exact wording.

INSTRUCTION_FOLLOWING score (0-5):
- 5: Every requested change exists and is exactly correct; nothing missing/misapplied; no extra edits.
- 4: All requested changes exist and are mostly correct; only a tiny inaccuracy.
- 3: Most requested changes exist but at least one is incomplete/incorrect/missing detail.
- 2: Only some requested changes exist; notable misses.
- 1: Requested changes largely not performed or substantially incorrect.
- 0: Contradicts or ignores the instruction entirely.

Output a single JSON object with:
- instruction_following_score (0-5)
- instruction_following_reason (one sentence, specific evidence)
\end{lstlisting}

\begin{lstlisting}[style=promptstyle,caption={IF judge --- user prompt.},label={lst:if-user}]
--- USER INSTRUCTION (what the model received) ---
{instruction}

--- AGENT'S EDITS (INITIAL state -> PREDICTION) ---
{formatted_diff}

CRITICAL JUDGMENT INSTRUCTIONS:
- Did the agent perform the changes the Instruction requires?
- Are the agent's edits scoped only to what the instruction asks (no unrelated edits)?
- If the diff is empty/trivial while the instruction clearly requires changes -> Low score (1 or 0).
- If the agent's edits implement the requested changes precisely -> High score (4-5).
- If the instruction targets a structured visual (table/chart/list/diagram), edits to the nested FRAMEs/RECTANGLEs/TEXTs ARE those edits, not unrelated edits.

REMINDER: Judge if the agent's edits ACHIEVED THE SEMANTIC INTENT of the Instruction.
\end{lstlisting}

\end{document}